\PassOptionsToPackage{table}{xcolor}
\documentclass{article}

\usepackage[main, final]{neurips_2026}

\usepackage[utf8]{inputenc} %
\usepackage[T1]{fontenc}    %
\usepackage{hyperref}       %
\usepackage{url}            %
\usepackage{booktabs}       %
\usepackage{amsfonts}       %
\usepackage{nicefrac}       %
\usepackage{microtype}      %
\usepackage{xcolor}         %
\usepackage{graphicx}
\usepackage{subcaption}
 \usepackage{pgfplots}
\pgfplotsset{compat=1.18}

\usepackage{graphicx} %
\usepackage{times}
\usepackage{latexsym}
\usepackage{amsmath}
\usepackage{amsthm}
\usepackage{booktabs}
\usepackage{multirow}

\usepackage{hyperref}
\usepackage{url}

\usepackage{multicol}

\usepackage[table]{xcolor}
\usepackage{colortbl}

\usepackage{array}
\usepackage{makecell}
\usepackage{siunitx}
\usepackage{tabularx}

\usepackage{pgfplots}
\pgfplotsset{compat=1.18}
\usepackage{rotating}
\usepackage{natbib}
\usepackage{algorithm}
\usepackage{algpseudocode}

\usepackage[most]{tcolorbox}
\usepackage{subcaption}
\usepackage{pifont}
\usepackage[normalem]{ulem}
\usepackage{caption}
\usepackage{amssymb}

\theoremstyle{plain}

\theoremstyle{definition}

\theoremstyle{remark}

\definecolor{mutedblue}{RGB}{76,114,176}
\definecolor{mutedorange}{RGB}{221,132,82}
\definecolor{mutedgreen}{RGB}{85,168,104}
\definecolor{mutedred}{RGB}{196,78,82}
\definecolor{mutedpurple}{RGB}{129,114,179}
\usepackage[most]{tcolorbox}

\usepackage{amsmath}
\usepackage{amssymb}
\usepackage{mathtools}
\newtheorem{Definition}{Definition}
\newtheorem{Proposition}{Proposition}

\newcommand{\method}{\textsc{Tale }}
\usepackage[T1]{fontenc}
\DeclareFontShape{T1}{ptm}{m}{scit}{<->ssub*ptm/m/sc}{}
\usepackage[utf8]{inputenc}
\usepackage{microtype}
\usepackage{inconsolata}
\providecommand{\hidden}[1]{}
\usepackage{graphicx}

\author{%
  Krish Sharma\textsuperscript{*} \\
  Université de Toulouse
  \And
  Omar Naim\textsuperscript{*} \\
  Université de Toulouse
  \AND
  Soumadeep Saha \\
  Université de Toulouse
  \And
  Vinija Jain \\
  Meta
  \And
  Aman Chadha \\
  Apple
  \And
  Nicholas Asher \\
  CNRS
  \AND
  {\normalfont\small \textbf{\textsuperscript{*}Equal contribution (co-first authors); listed alphabetically by first name.}}
}

\title{
TAPIOCA: Why \underline{T}ask-\underline{A}ware \underline{P}runing 
\underline{I}mproves \underline{O}OD model \underline{Ca}pability
}
\begin{document}

\maketitle

\begin{abstract}

Recent work \citep{peer:etal:2022,tale} has shown that task-aware layer pruning can improve model performance on particular tasks. We investigate when and why such improvements occur, using small transformers trained from scratch as testbeds on controlled polynomial regression tasks, then confirming our findings on large language models (LLMs).  In both cases, such pruning yields no increase in accuracy on in-distribution (ID) data but consistently improves out-of-distribution (OOD) performance and generalizes such gains across multiple distribution shifts. We offer a geometric explanation of this phenomenon. Each task with an ID distribution induces a task-adapted geometry. OOD inputs induce layerwise norm, pairwise-distance and variance structure profiles that can deviate from the corresponding ID profiles. Task-aware pruning identifies layers that amplify this distortion; removing these layers shifts OOD profiles toward those of the adapted distribution. We provide causal evidence that this geometric realignment improves OOD performance through controlled distribution shifts and residual-scaling interventions. This behavior is consistent across model scales.

\end{abstract}

\section{Introduction}

Recent work has promoted task-aware pruning: \emph{removing layers from large language models (LLMs) at inference time can improve task performance}, even without retraining \citep{peer:etal:2022,tale}. We analyze this surprising phenomenon using \citep{tale}'s method \method , which reports the largest gains. Our analysis surprisingly reveals that \method{} improves performance on out-of-distribution (OOD) inputs, but not on inputs that align with the model's in-distribution (ID) training data. We use geometric statistics  \citep{hosseini:fedorenko:2023} to capture structural changes in representation spaces, and find that OOD inputs induce layer-wise \textbf{representational norms, pairwise distances, and variance shifts} that distort those induced by ID inputs.  Distribution shift thus not only perturbs the input; it progressively changes the geometry of internal representations.

This suggests a distribution-dependent interpretation of a transformer layer's function: \emph{layer importance is conditional not only on the task, as task-aware pruning methods have argued, but also on the input distribution}. Layers that are beneficial for ID precision can become harmful under distribution shift. By deleting layers that amplify distortions in representational geometry, task-aware layer pruning makes OOD representations better fit the geometry learned from the adapted distribution.

Establishing these results requires a setting with explicit distinctions between ID and OOD data. We thus begin with the in-context learning framework of \citet{garg:etal:2022}, using a controlled setting that precisely specifies both the task and the data distribution. This gives us a diagnostic tool for uncovering the mechanisms underlying \method. We then extend the analysis to pretrained LLMs and show that \textbf{the same ID/OOD asymmetry holds at scale}, indicating that the phenomenon is not an artifact of small models.  More specifically, our paper makes the following contributions:

  \begin{figure}[!ht]
    \includegraphics[width=\linewidth]{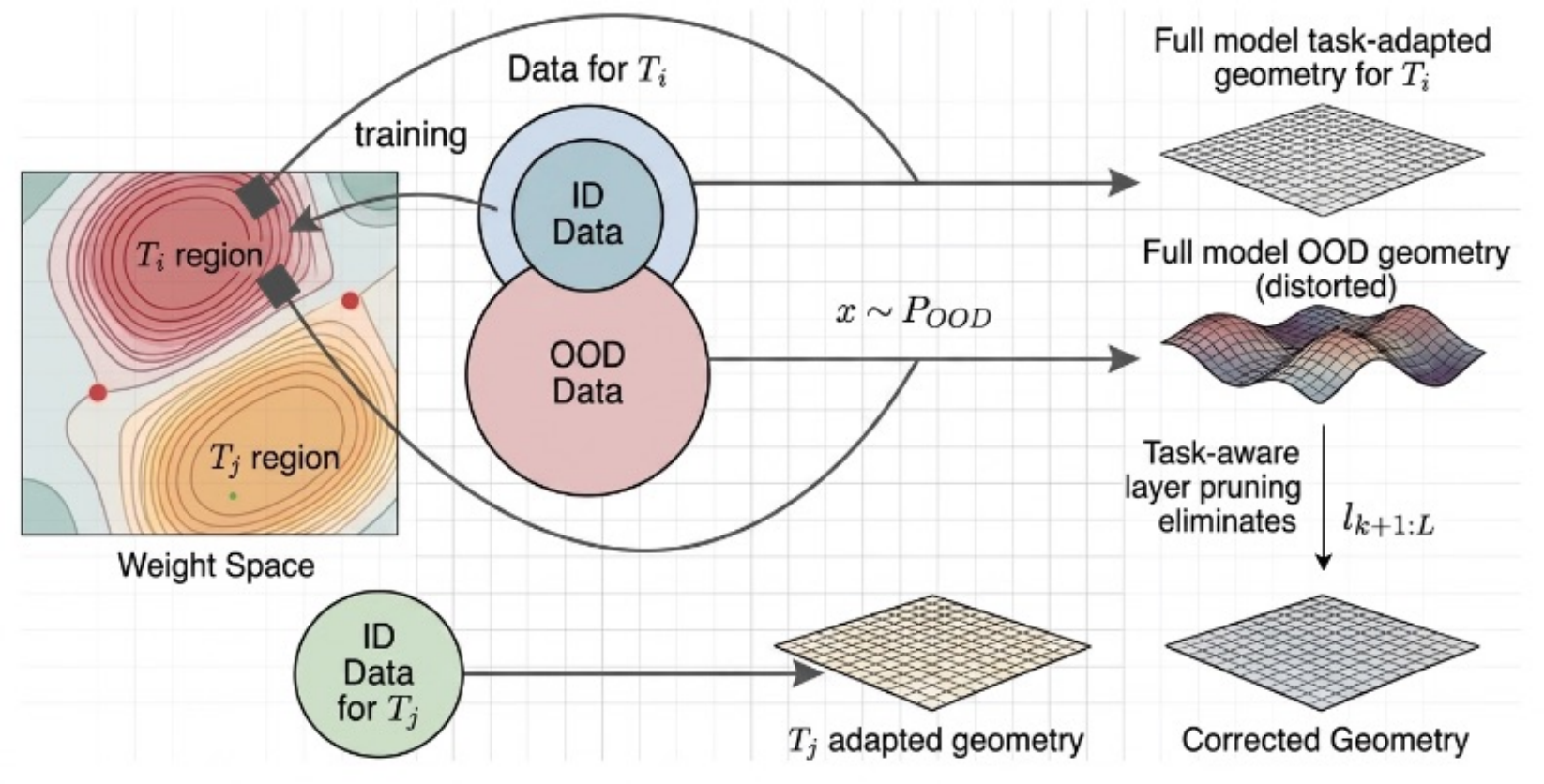}
    \caption{Weight-space-defined functions for different tasks ($T_i, T_j$), applied to data from different distributions (ID, OOD), characterize task-specific geometries; ID data for $T_i$ give rise to a task-adapted geometry; OOD data give rise to a distorted geometry, which task-aware layer pruning can correct.} 
    \label{fig:manifold_projection}
\end{figure}

\noindent
\textbf{1. Task-aware pruning gains are OOD-specific.}
\method{} does not improve performance when evaluated on the (ID) distribution to which the model is adapted. In contrast, \method{} consistently improves performance under distribution shift. This pattern holds both for small transformers trained on controlled regression tasks and for larger LLMs evaluated on shifted NLP benchmarks.

\noindent
\textbf{2. OOD inputs induce a mismatch in representation geometry.}
For a given task, ID inputs induce characteristic layerwise hidden-state norms, pairwise distances, and variance structure that define the task's \emph{adapted representation geometry}. OOD inputs can move hidden states away from this geometry, producing geometric profiles that differ from those observed on ID data. The degree of this mismatch empirically tracks distribution shift and performance degradation. Pruning layers selected by \method{} reduces the mismatch, moving OOD profiles toward the ID profile.

\noindent
\textbf{3. Some layers act as distribution-dependent amplifiers.}
We propose a local and distribution-dependent view of transformer layers: the same layer can act as a useful refinement on ID inputs and as a harmful amplifier on OOD inputs.  Local linear surrogate analyses show that certain layers exhibit high-gain, low-rank amplification on far-OOD inputs while behaving more benignly on near-ID inputs.  These surrogates help provide an argument that geometric information causally affects performance.

\noindent
\textbf{4. A geometric account of why pruning helps.}
A model's training associates each task with a local representation geometry. ID inputs fit this geometry; OOD inputs may not. Pruning improves OOD performance by removing layers that amplify the mismatch between OOD and ID geometries. On ID data, however, these layers contribute to the task geometry, so pruning does not help. This explains why task-aware pruning is task-specific, why it helps under distribution shift but not on ID inputs: its effect is to reduce geometric mismatch, whether by contraction or expansion.

\paragraph{Outline and Scope of this work.}
In what follows, after a literature review, Section~\ref{sec:ood} establishes the OOD-specificity of \method{} gains in both small and large models. Section~\ref{sec:analysis} presents our geometric analysis of this OOD-specificity.  We focus on \method{} because it is training-free and has larger reported task-specific gains than alternative pruning methods \citep{tale,peer:etal:2022,sleb,zhong:2025}.  We aim, however, not to compare pruning methods or OOD mitigation strategies, but to explain representational mechanisms underlying task-aware pruning gains.

\section{Related Work}
\label{sec:related}

\paragraph{From compression to task-aware pruning.}
Layer pruning framed as compression should reduce computational cost while preserving a general-purpose quality signal, often with some accuracy degradation. \citep{sleb,zhong:2025} remove blocks by representation similarity or perplexity; \citep{slicegpt} reduce dimensionality; \citep{sparsegpt,wanda} apply unstructured pruning based on reconstruction error, weight magnitude, or activations. These methods are largely task-agnostic; improving task accuracy is not their primary objective. In contrast, task-aware pruning seeks to remove layers to \emph{improve} task performance. Building on \citet{peer:etal:2022}, \method{} reports accuracy gains across several LLM families by optimizing task-specific validation accuracy \citep{tale}. We study when and why such gains occur.

\paragraph{Layer importance: task- and distribution-dependent.}
Prior work disagrees on which layers matter most. \citet{dalvi:etalL:2020} find early layers to be critical via probing and similarity analysis, while \citep{ellie,latestlayers1,latestlayers2} emphasize later layers as the locus of task-relevant representations. \method{} challenges both views: layer importance is neither purely positional nor fixed, but task- and distribution-dependent \citep{tale}.  In support, we show that the same layer can act as a useful refinement on in-distribution inputs and as a low-rank, high-gain amplifier under distribution shift, thereby distorting representations and degrading performance.

\paragraph{Representational geometry and layer analysis.}
\citet{hosseini:fedorenko:2023} find that LLMs learn to straighten token trajectories across depth, reducing representational curvature. \citet{skean:etal:2025} identify a compression--reconstruction pattern through matrix entropy. \citet{zhang:etal:2024} analyze layer redundancy through representation similarity. The information-bottleneck perspective \citep{tishby2015deep} predicts that layers selectively compress task-relevant information. We build on this geometric framing, but identify a different organizing principle: whether a layer helps or hurts depends on how it transforms representations under distribution shift.  

Harmful layers can exhibit \emph{anisotropic amplification}, increasing representational norms, pairwise distances and shift variance structure for OOD inputs, pushing them away from the geometry given by the adapted distribution. In contrast, beneficial layers perform transformations improving in-distribution performance. Appendix~\ref{appendix:alternatives} shows that trajectory linearity, entropy compression, and information-bottleneck effects alone do not predict \method{}'s gains. The key missing ingredient is a \emph{distribution-dependent view of layer behavior}: layers do not have fixed utility, but interact with the input distribution to either preserve or distort the model's learned representation geometry.  Our controlled regression setting \citep{garg:etal:2022} with models trained from scratch enables a precise control over the input distribution not possible with large pretrained models.

\hidden{}

\hidden{}

\section{\method{} Improves OOD, Not ID Performance}
\label{sec:ood}

To determine when task-aware layer elimination improves performance, we need a clear distinction between in-distribution (ID) and out-of-distribution (OOD) evaluation data. For standard pretrained language models, this distinction is difficult to define because the training distribution is unknown and probably heterogeneous. So we begin with a controlled in-context regression setting, where the training distribution is explicit, and then test whether the same pattern appears in fine-tuned large language models. Across both settings, \method{} does not improve performance when the evaluation distribution matches the distribution on which the model is trained or fine-tuned. In contrast, under distribution shift, \method{} consistently identifies layers whose removal improves accuracy.

\subsection{\method{} in a controlled setting: In-context Linear Regression}
\label{sec:ood-regression}
We train a 12-layer, 8-head transformer, $Ba_1$, to perform in-context linear regression. Each prompt contains context pairs $(x_i,y_i)$ generated by a single linear function\footnote{The code is available at: \url{https://github.com/omyokun/TAPIOCA}}
\[
f(x) = ax + b, \qquad a,b \sim \mathcal{U}(-\sigma,\sigma).
\]
Inputs $x$ are sampled independently from $\mathcal{U}(-1,1)$. Given the context pairs and a query input $x_q$, the model outputs a prediction for $f(x_q)$ without any gradient updates.

During training, we set $\sigma=1$. Following the same evaluation protocol as in~\cite{naim-etal-2026-ssa}, we evaluate $Ba_1$ on both ID data ($\sigma=1$) and OOD data ($\sigma>1$). We apply \method{} to $Ba_1$ separately for each validation distribution to obtain the best-performing model found by pruning at each scale, denoted by $Be_1, Be_2, \ldots, Be_\sigma, \ldots$.\footnote{At each step, \method{} removes the layer whose deletion yields the largest improvement in the validation metric. It stops when no further deletion improves that metric.}

\paragraph{\method{} does not improve on ID validation data.}
When \method{} uses ID validation data from $\mathcal{U}(-1,1)$, it removes no layers without loss of accuracy. Thus, $Ba_1 = Be_1$ is optimal for this distribution, and $Ba_1$'s layers are needed for ID prediction. When $Be_\sigma$ for $\sigma > 1$ is evaluated on $\mathcal{U}(-1,1)$, it performs substantially worse than $Ba_1$: $Be_2$ has mean MSE $0.011770$, compared with $0.000008$ for $Ba_1$. Thus, pruning does not produce a uniformly better regressor. Instead, it sacrifices ID precision while improving performance under distribution shift.

\paragraph{\method{} improves performance under distribution shift.}

\begin{figure}[!h]
    \centering
    \includegraphics[width=\linewidth]{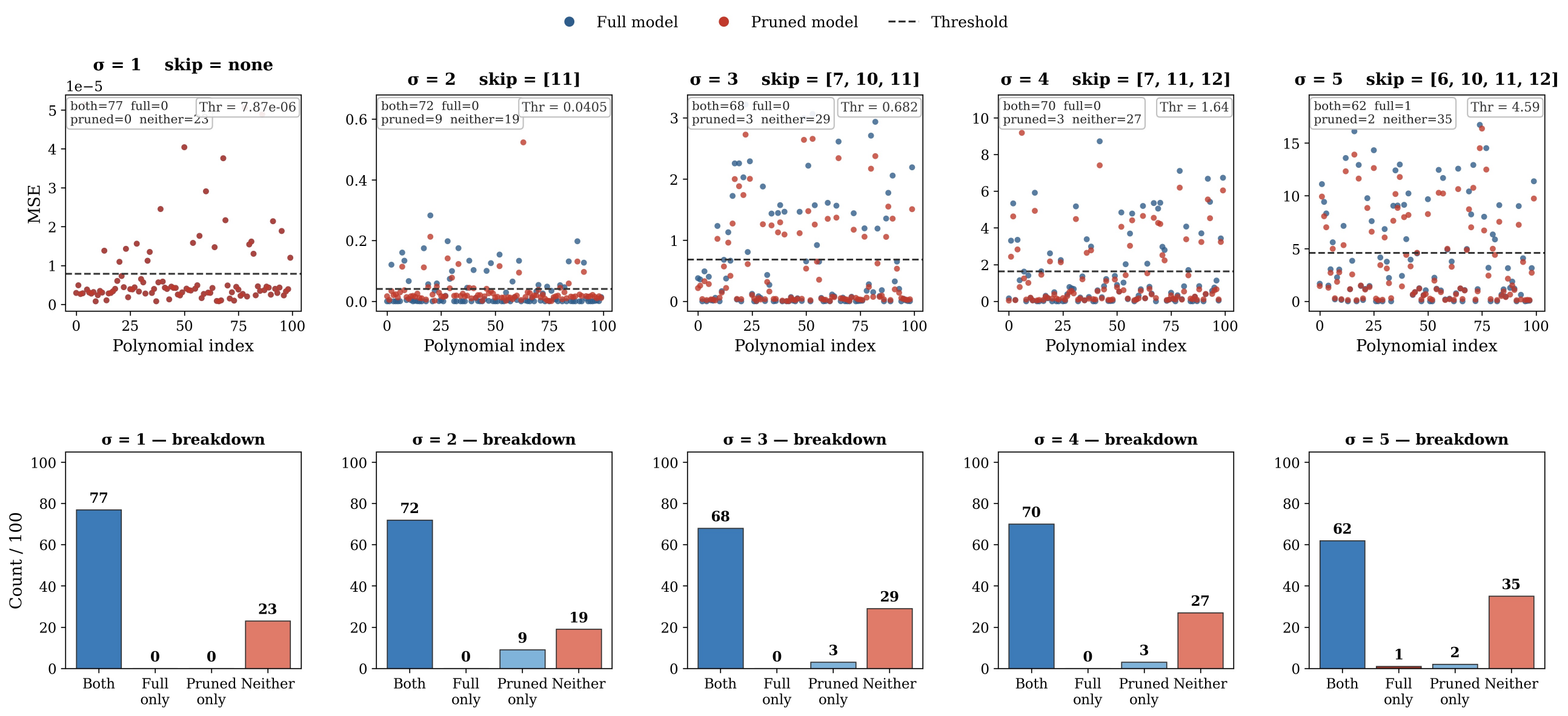}
    \caption{Threshold analyses over linear functions sampled from $U(-\sigma,\sigma)$. The dotted line denotes mean base-model MSE for $1 \leq \sigma \leq 6$. The pruned model $Be_2$ sacrifices strong in-distribution performance, including nearly $1500\times$ worse mean MSE on $U(-1,1)$, while systematically outperforming the base model $Ba_1$ on OOD samples from $U(-2,2)$, where mean MSE drops from $0.04054$ to $0.03440$. The plots show that pruning selectively benefits functions outside the training geometry rather than uniformly improving regression. This is evidence that dropped layers are necessary for optimal in-distribution precision but become liabilities under distribution shift.}
    \label{fig:threshold}
\end{figure}

As illustrated in Figure~\ref{fig:threshold}, the behavior changes on OOD data. As the coefficient range widens beyond the training distribution, \method{} begins to identify removable layers. On $\mathcal{U}(-2,2)$, for example, the full model obtains mean MSE $0.04054$, while the pruned model obtains mean MSE $0.03440$. We report additional results in Table~\ref{linear2} of Appendix~\ref{appendix:regression}. This improvement is not uniform across all functions. A per-function analysis shows that pruning primarily benefits functions outside the training geometry, while harming or failing to improve functions close to the training distribution. Thus, \method{} does not improve the model unconditionally. It improves performance specifically on shifted inputs, where the full model's learned transformations are no longer well calibrated.

\subsection{Large-model setting: fine-tuned math and code experts at two scales}
\label{sec:math-experts}

The same pattern appears in pretrained LLMs across model scales. Because the pretraining distribution of instruction-tuned LLMs is not directly observable, standard benchmarks do not provide a clean ID/OOD split. To obtain a clearer split, we fine-tune models on two task families: mathematical reasoning and code. We illustrate the setup with the math specialists.

We treat mathematical reasoning as the in-distribution task for each math-specialized model: \emph{(i)} Llama 3.1 8B, fine-tuned on NuminaMath-CoT, denoted $M_{\text{math}}^{8B}$ and evaluated on MATH500 as the ID test; and \emph{(ii)} GPT-OSS 120B, fine-tuned on the same mathematical reasoning corpus, denoted $M_{\text{math}}^{120B}$ and evaluated on MMLU-Math\footnote{For the larger-scale fine-tuned model, we use MMLU-Math as the in-distribution proxy because it is the harness-supported math evaluation closest to the model's fine-tuning distribution.} as the ID test. In both cases, we then evaluate on full MMLU and BoolQ as OOD tasks relative to the model's math specialization. We construct analogous code specialists and OOD domains for them; training details are provided in Appendix~\ref{appendix:training-llms}.

The pattern is consistent across both scales and task families (Table~\ref{tab:tale-scales}). At 8B, the math-specialized model achieves $87.5\%$ on MATH500, and \method{} cannot remove any layer without reducing performance below baseline. On MMLU, the same model scores $36.7\%$ at baseline. \method{} removes layers and reaches $44.1\%$, an absolute gain of $7.4$ points, corresponding to a $20\%$ relative gain. At 120B, the math-specialized model achieves $94.0\%$ on the ID task, and \method{} again cannot prune any layer without loss of accuracy. On full MMLU, however, \method{} yields a $+7.0$ point absolute gain.

\begin{table}[h]
\centering
\small
\begin{tabular}{llccc}
\toprule
Model & Task & Baseline & \method{} & $\Delta$ \\
\midrule

\multirow{3}{*}{$M_{\text{math}}^{8B}$ (Llama-3.1)}
& In-dist (MATH500) 
& $87.5 \pm 0.3$ & $87.5 \pm 0.2$ ($\emptyset$) & $0.0$ \\
& OOD (MMLU Math) 
& $36.7 \pm 0.5$ & $44.1 \pm 0.4$ & $+7.4$ \\
& OOD (BoolQ) 
& $83.7 \pm 0.4$ & $86.7 \pm 0.3$ & $+3.0$ \\

\midrule
\multirow{4}{*}{$M_{\text{CS}}^{8B}$ (Llama-3.1)}
& In-dist (Code Alpaca) 
& $55.0 \pm 0.6$ & $51.0 \pm 0.7$ ($\emptyset$) & $-4.0$ \\
& OOD (MMLU Math) 
& $43.0 \pm 0.5$ & $45.4 \pm 0.4$ & $+2.4$ \\
& OOD (MMLU high school CS) 
& $74.0 \pm 0.5$ & $78.0 \pm 0.4$ & $+4.0$ \\
& OOD (MMLU college CS) 
& $53.0 \pm 0.6$ & $61.0 \pm 0.5$ & $+8.0$ \\

\midrule
\multirow{3}{*}{$M_{\text{math}}^{120B}$ (GPT-OSS)}
& In-dist (MATH500) 
& $94.0 \pm 0.2$ & $94.0 \pm 0.2$ ($\emptyset$) & $0.0$ \\
& OOD (MMLU Math) 
& $47.0 \pm 0.4$ & $54.0 \pm 0.3$ & $+7.0$ \\
& OOD (BoolQ) 
& $93.0 \pm 0.3$ & $94.0 \pm 0.2$ & $+1.0$ \\

\midrule
\multirow{4}{*}{$M_{\text{CS}}^{120B}$ (GPT-OSS)}
& In-dist (Code Alpaca) 
& $81.0 \pm 0.4$ & $78.0 \pm 0.5$ ($\emptyset$) & $-3.0$ \\
& OOD (MMLU Math) 
& $41.0 \pm 0.5$ & $44.6 \pm 0.4$ & $+3.6$ \\
& OOD (MMLU high school CS) 
& $85.0 \pm 0.3$ & $88.0 \pm 0.3$ & $+3.0$ \\
& OOD (MMLU college CS) 
& $78.0 \pm 0.4$ & $81.0 \pm 0.4$ & $+3.0$ \\

\bottomrule
\end{tabular}
\vspace{0.5em}
\caption{\method{} applied to two fine-tuned math specialists at different scales and two fine-tuned code specialists. Results are reported as mean accuracy across random seeds with standard deviation. The symbol $\emptyset$ indicates that no layer deletion improved ID performance. On OOD tasks, \method{} improves performance across task families, scales, and architectures.  All evaluations were done in LM-eval.}
\label{tab:tale-scales}
\end{table}

{\bf Takeaways.} The large-model experiments span a $15\times$ difference in parameter count and two architecture families: a dense transformer at 8B and a sparse mixture-of-experts model at 120B. They nevertheless produce the pattern seen in the controlled regression setting: \method{} cannot improve the model on its adapted task, but substantially improves performance on tasks that are OOD relative to that adaptation. The removed layers are not globally unnecessary. They are useful for the distribution on which the model is adapted.  But they can become harmful under distribution shift. This indicates that the effect is not an artifact of small transformers, a single model family, or a particular fine-tuning procedure. Instead, the utility of a layer depends on the input distribution. Section~\ref{sec:analysis} analyzes the representational mechanism behind this effect.

\subsection{ Generalizing OOD performance} Accuracy increases on OOD are not specific to just one targeted distribution.  They generalize across distribution shifts.  We performed a test where we TALE-optimized a model on distribution A and then tested the optimized model on distribution B. Using synthetic data and task from Section 3.1, we optimized a model on $\mathcal{U}(-2,2)$ and $\mathcal{U}(-3,3)$ (using two different optimizers), which eliminates 4 to 6 layers to improve performance on the targeted distribution.  But the pruned models also performed better on $\mathcal{U}(-4,4)$ through $\mathcal{U}(-10,10)$---between $0.82$ to $0.93$ of the baseline's MSE. We did a similar experiment on larger models and NLP benchmark datasets. We used the baseline Llama model that was optimized with \method{} on MMLU high school mathematics. While the base model got 36.7\%, the \method{}-optimized model got 44.1\% accuracy. We then took that model \method{} optimized for high school math and tested it as is on the MMLU High School Computer Science dataset. The math \method-optimized model increased the accuracy on the test set to 68\% over the base model's score of 66\%. Using \method{} to optimize a model for one distribution offers accuracy improvements with other distribution shifts and thus generalizes beyond targeted effects.

\section{Analysis: Pruning Aligns OOD Representation Geometry}
\label{sec:analysis}

Section~\ref{sec:ood} showed that \method{} improves performance primarily under distribution shift. We now investigate why. OOD inputs induce distorted representation geometry inside the network: their hidden states exhibit norms and pairwise distances that differ systematically from those induced by ID inputs. \method{} improves OOD performance by removing layers that amplify this distortion, thereby moving OOD representations closer to the geometry induced by the model's adapted distribution. Formal details are provided in Appendix~\ref{appendix:math}.

 Three experiments further support the geometric explanation. First, in the controlled regression setting, we show that OOD inputs produce layer-wise distance profiles that differ systematically from ID inputs, and that \method{} shifts the OOD profile toward the ID profile. We replicate this effect in a fine-tuned Llama 3.1 8B model: pruning improves MMLU accuracy while contracting MMLU representations toward the MATH500 profile. Third, we show at least for small models that OOD inputs have a significantly different variance structure from ID inputs and that pruning brings OOD variance structure back more in line with ID variance structure.  Fourth, we provide causal and layer-level evidence by rescaling residual updates and fitting linear surrogates to individual layers and introducing auxiliary ‘geometry-fixing’ layers that realign representations for OOD inputs and restores performance.

\subsection{Norm and distance shifts}
\label{sec:llm_geometry}

\paragraph{OOD inputs induce distorted distance profiles.}
In the controlled regression setting of Section~\ref{sec:ood}, we can directly compare representations induced by ID and OOD inputs. For each 200 total prompts, we extract hidden states at every transformer layer. We focus on the final query token and compute its distance to preceding token representations. Averaging across prompts gives a layerwise distance profile for each evaluation distribution. This profile reveals how the model separates token representations across layers.

\begin{figure}[t]
    \centering
    \includegraphics[width=\linewidth]{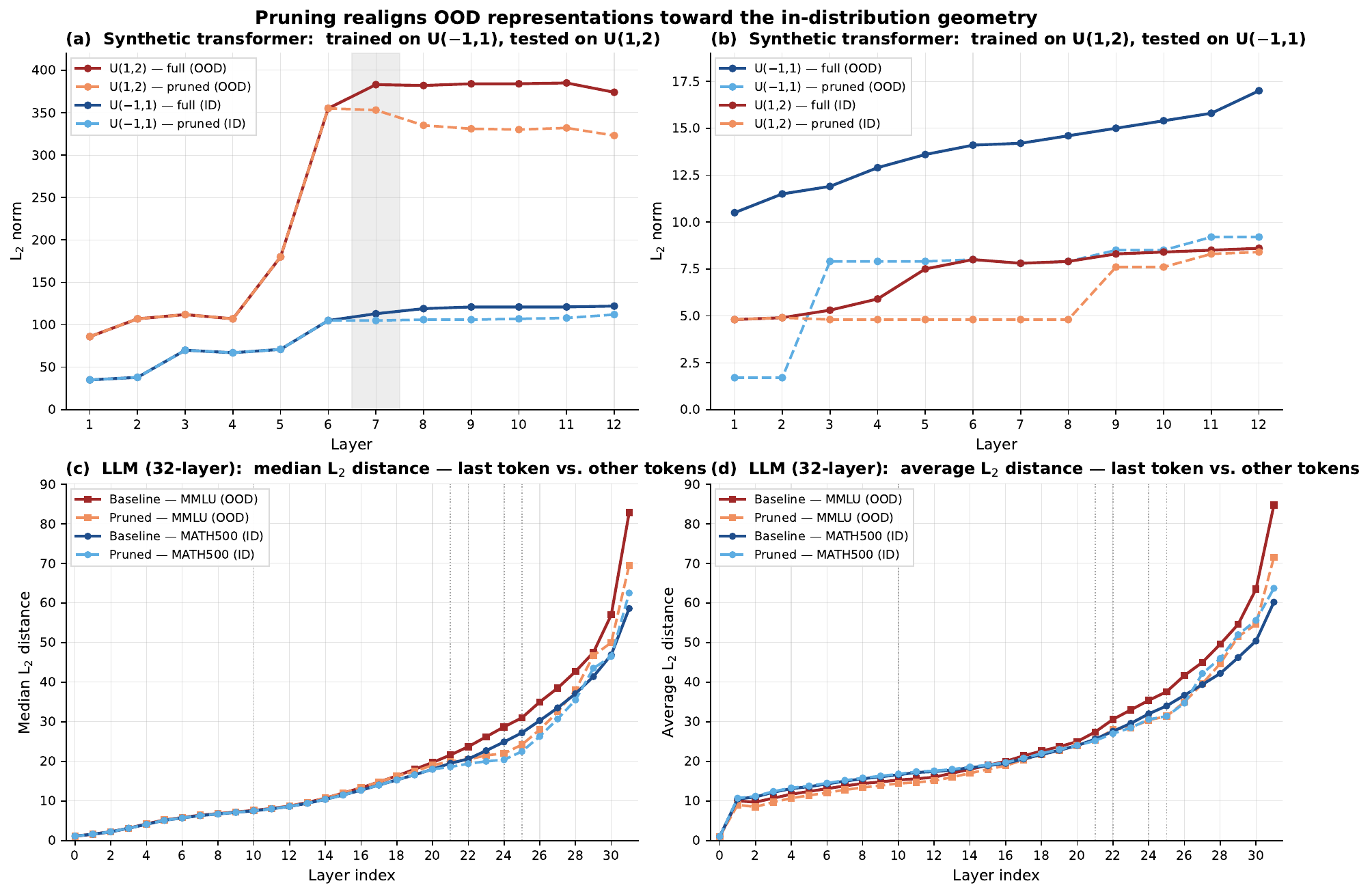}
    \caption{\textbf{Pruning realigns OOD representations toward the in-distribution geometry.}
    \emph{Top: regression-task results for $L_2$ median distance from the final token to prior tokens.}
    \textbf{(a)} The model is trained on $U(-1,1)$ and tested on $U(1,2)$: OOD distances inflate to ${\sim}385$, and \method{} contracts them toward the ID trajectory.
    \textbf{(b)} With train/test roles reversed, OOD with lower valued inputs still produce a higher norm than ID; thus the higher OOD norm is not just a function of higher input values.  Pruning shrinks OOD L2 distances toward the ID baseline, showing that \method{} matches the task-specific geometry rather than merely suppressing activations.  \emph{Bottom:} Llama 8B $L_2$ distances from the final token to preceding tokens, using 200 MATH500 and 200 MMLU prompts.
        \textbf{(c)} Median and \textbf{(d)} average distances diverge between MMLU (OOD) and MATH500 (ID) after layer~14; removing layers $\{10,21,22,24,25\}$ contracts the OOD trajectory toward the ID baseline and improves MMLU accuracy by $+7.4$ points. Figure~\ref{fig:math-mmlu-l1} shows the same pattern with $L_1$ distances.}
    \label{fig:pruning_geometry}
\end{figure}

For the base model $Ba_1$, OOD inputs sampled from the coefficient range $\mathcal{U}(1,2)$ produce larger hidden-state distances than ID inputs. As shown in Figure~\ref{fig:pruning_geometry}(a), this increase is not confined to the input representation; it grows across intermediate layers, indicating that distribution shift is amplified by the network. Thus, OOD degradation is associated with a change in the internal geometry of the residual stream, not merely with larger inputs at the output layer. Applying \method{} to the OOD task produces a pruned model whose distance profile contracts toward the ID profile, as shown by the dotted line in Figure~\ref{fig:pruning_geometry}(a). Pruning changes the trajectory of OOD representations through the network, making them more similar to the representations on which the model was trained to perform well. On the ID task, however, the same pruning operation worsens performance.
\paragraph{\method{} aligns geometry rather than suppressing norms.}
\label{sec:alignment_not_suppression}
Figure~\ref{fig:pruning_geometry}(a) might suggest that \method{} improves OOD performance simply by reducing inflated representation distances; that is, pruning might merely suppress activation norms. Figure ~\ref{fig:pruning_geometry}(b), however, where we test a model on coefficients sampled from $\mathcal{U}(1,2)$ and evaluate it on data sampled from $\mathcal{U}(-1,1)$, rules out this explanation.

In this setting, the OOD distribution does not induce the same direction of distance inflation as before. If \method{} merely suppressed norms, pruning should still reduce the distance profile. Instead, pruning \textbf{increases} representation distances. The consistent effect across both directions of shift is therefore \textbf{not contraction, but alignment}: \method{} moves the OOD distance profile toward the profile induced by the model's training distribution. This shows that \method{} does not simply remove high-norm layers or act as a regularizer. Its effect depends on the relation between the geometry induced by the OOD distribution and the ID-adapted task geometry.

\paragraph{The same geometry shift appears in a fine-tuned LLM.}  We use the fine-tuned math model $M_{\mathrm{math}}$ from Section~\ref{sec:ood}. We treat MATH500 as its adapted task and MMLU as OOD relative to this specialization. Appendix~\ref{appendix:nlp-benchmarks-norm} provides additional benchmark-level evidence that most standard NLP benchmarks do not behave like ID data for this model.

For each of 200 prompts and each transformer layer, we extract the hidden state of the final token and compute the median distance to the preceding token states. Averaging over prompts gives a layer-wise distance profile for each dataset. In the full model, the MMLU profile diverges from the MATH500 profile in the middle and late layers, as shown in Figure~\ref{fig:pruning_geometry}(c,d). From this point on, OOD representations have larger pairwise distances than ID representations, and the gap persists through the final layer.

Removing the TALE-selected layers has an asymmetric effect. On MATH500, pruning perturbs the representation profile and the geometric statistics and degrades performance---implying these layers are useful for the adapted task. On MMLU, the same pruning operation substantially contracts the distance profile toward the MATH500 profile and improves accuracy by $7.4$ points. Thus, the large-model result mirrors the controlled regression result: pruning helps when it corrects an OOD representation trajectory, but hurts when it disrupts an already well-adapted ID trajectory.

\method{} thus can improve benchmark performance without contradicting the usefulness of depth. Removed layers are not intrinsically redundant. They are useful under one distributional regime and harmful under another. Their contribution depends on whether the input follows the geometry for which the model's transformations are calibrated.

 \subsection{Covariance geometry shifts}
We now turn to covariance and spectral differences.  For the regression models, we were able to estimate covariance without difficulty.  We used a small 12 layer regression task model trained on inputs $x \sim \mathcal{U}(0,1)$ (ID). OOD inputs are
drawn from $\mathcal{U}(-1,0)$, a disjoint interval of equal magnitude, so that
the shift changes the input \emph{distribution} without changing its scale.  

For a given prompt, we track the representation of the 
final input token $X$, whose hidden state produces the model's prediction, as it
passes through the network from its input embedding $X_0$ to the final layer representation $X_L$.  This gives the trajectory,
    $X_0 \;\to\; X_1 \;\to\; \cdots \;\to\; X_{L}$.  
At each layer $\ell$ we collect the query-token representations of $N$ ID samples
and $N$ OOD samples, giving matrices in $\mathbb{R}^{N\times d}$ ($d$ is the embedding dimension), and
form their empirical covariances $\Sigma^{\text{ID}}_\ell$ and
$\Sigma^{\text{OOD}}_\ell$. A small trace-shrinkage,
$\Sigma \leftarrow (1-\alpha)\Sigma + \alpha\,\tfrac{\operatorname{tr}\Sigma}{d} I$
with $\alpha=10^{-2}$, ensures the covariances are invertible.

We compare ID and OOD covariance via  \emph{relative covariance}
\[
    M \;=\; \Sigma_{id}^{-1/2}\,\Sigma_{ood}\,(\Sigma_{id}^{-1/2})^\top,
\]

$\Sigma_{ood}$ in the coordinate frame in which $\Sigma_{id}$ is the
identity. Its eigenvalues $\{\mu_j\}$ jointly capture differences in both the
\emph{scale} and the \emph{orientation} of the two representation clouds:
$\mu_j = 1$ for all $j$ if and only if the two covariances coincide, while
$\mu_j > 1$ (resp.\ $<1$) marks a direction in which $\Sigma_2$ carries more
(resp.\ less) variance than $\Sigma_1$. We summarize the deviation from identity with the measure over the two positive definite matrices:
\[
    S(\Sigma_1,\Sigma_2) \;=\; \frac{1}{d}\sum_{j=1}^{d}\bigl|\log \mu_j\bigr|,
\]

$S(\Sigma_1,\Sigma_2) = 0$ for $\Sigma_1 = \Sigma_2$; it grows symmetrically with over- and
under-inflation of any direction.

At each layer we evaluate two comparisons:
\begin{itemize}
    \item \textbf{ID--ID (baseline).} Two disjoint halves of the ID samples,
    $S(\Sigma^{\text{ID},1}_\ell, \Sigma^{\text{ID},2}_\ell)$. This measures the
    finite-sample noise floor: how far from identity two covariances of the
    \emph{same} distribution fall. 
    \item \textbf{ID--OOD (signal).} $S(\Sigma^{\text{ID}}_\ell,
    \Sigma^{\text{OOD}}_\ell)$. Deviation of this curve \emph{above} the baseline
    quantifies how differently the model represents OOD inputs.
\end{itemize}

\paragraph{Results} 
We find a stable covariance for the ID-ID baseline (negligible standard deviation from the mean) but a significant deviation of OOD from ID on all layers after the pruned layers.   Where $\widetilde{\Sigma}_{id/ood}$ are the normalized matrices, we have $S(\widetilde{\Sigma}_{id,L},\widetilde{\Sigma}_{ood,L}) = 0.642$, but for pruned models $\pi$ with the model's first layer removed, we have 
\[S(\widetilde{\Sigma}_{id,L},\widetilde{\Sigma}_{ood,L}) = 0.521 \]
Thus, pruning to improve accuracy moves OOD covariance closer to ID covariance, while removing layers that degrade accuracy worsens the distortion.\footnote{Due to the lower number of samples and the high dimensionality of the larger models, our covariance estimator did not work as it should.} Full scores and details are in Appendix \ref{appendix:math}.2, and we strengthen our account there with a formal  connection between covariance shift and MSE.  %

\subsection{OOD distortion is layer- and distribution-dependent}
\label{sec:linear_surrogate}

We now examine what distinguishes layers that distort OOD representations. For each layer $\ell$ and dataset $D$, we fit the best linear surrogate to the layer's residual-stream input-output pairs:
\[
W_{\ell}^{D}
=
\arg\min_W
\sum_i \|W x_i - y_i\|_2^2 .
\]
Here $x_i$ is the residual-stream input to layer $\ell$, and $y_i$ is the corresponding output. This surrogate  approximates the layer's local action on the evaluated data distribution.%

We examine the singular-value spectrum of $W_{\ell}^{D}-I$, which measures how far the layer departs from a passthrough map and whether this departure is spread across many directions or concentrated in a few. We also measure the on-data norm gain, $\|y_i\|/\|x_i\|$, which captures how much the layer expands or contracts representations from that distribution.

\begin{figure}[t]
    \centering
    \includegraphics[width=0.7\linewidth]{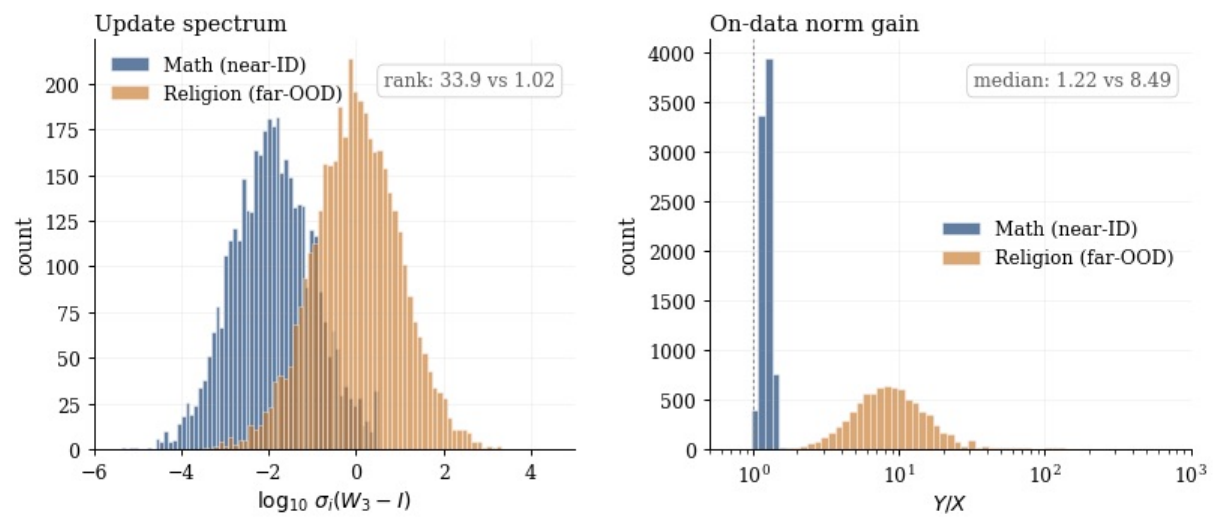}
    \caption{Distribution-dependence of the layer-3 linear surrogate $W_3$. We fit $W_3$ on Mathematics MMLU, a near-ID slice, and on Religion MMLU, a far-OOD slice. Left: distribution of $\log_{10}$ singular values of $W_3 - I$. Right: distribution of on-data norm gain $\|W_3 x\|/\|x\|$ across tokens, shown on a log scale. The same weights yield a diffuse, small-magnitude update on near-ID inputs, with stable rank~$33.9$ and median gain~$1.22$, but a near rank-one, order-of-magnitude amplification on far-OOD inputs, with stable rank~$1.02$, median gain~$8.49$, and a tail past $10^2$.}
    \label{fig:distribution_shift_surrogate}
\end{figure}

These diagnostics show that the same layer can behave very differently depending on the input distribution. In Figure~\ref{fig:distribution_shift_surrogate}, the surrogate for layer~3 has moderate gain and a diffuse update spectrum on the near-ID MMLU slice. On the far-OOD slice, the same layer exhibits much larger gain and a highly concentrated spectrum, indicating that it acts as a low-rank amplifier on those inputs. The weights are identical in both cases; only the input distribution changes.

This shows that harmful layer behavior is not an intrinsic property of a layer's weights alone. It is a joint property of the layer and the distribution of representations entering it. A layer that acts as a benign refinement on near-ID inputs can become a high-gain amplifier on far-OOD inputs. This provides a layer-level mechanism for the distance-profile results above: distribution shift can move representations into directions that certain layers amplify, and \method{} improves performance by removing or attenuating layers whose updates worsen this mismatch.

\subsection{Surrogate inverses provide causal evidence for geometric correction}
\label{sec:surrogates-causal}

\begin{figure}[t]
    \centering
    \includegraphics[width=9.5cm]{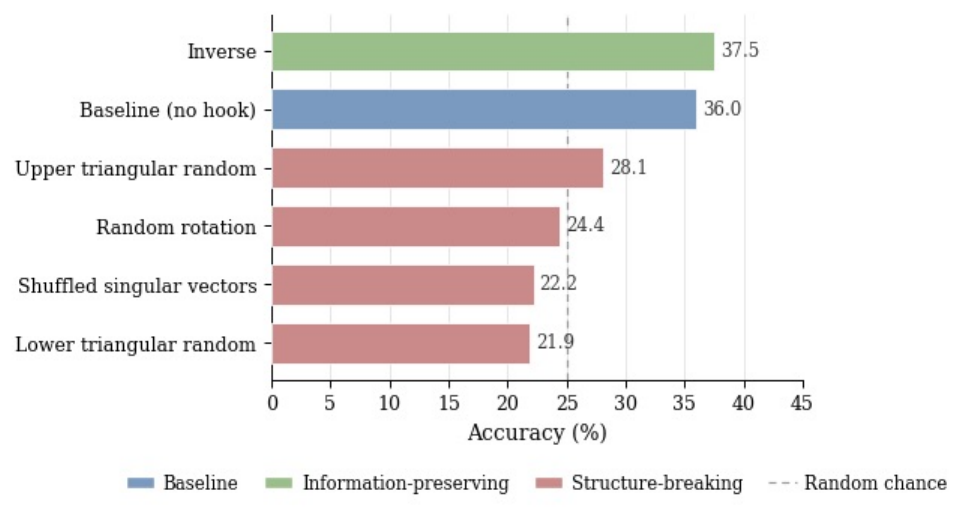}
    \caption{Accuracy on MMLU high-school mathematics, using 2-shot evaluation, under different residual-stream interventions applied at layer~24. The \emph{inverse} hook, which preserves task-relevant structure, matches or slightly exceeds the no-hook baseline, while random rotations and random triangular maps collapse performance to near the 4-way random-chance level, shown by the dashed line. All interventions preserve activation norms to within ${\sim}10\%$ ($\lVert M y \rVert / \lVert x \rVert \in [1.04, 1.14]$).}
    \label{fig:layer23-interventions}
\end{figure}

In prior sections we have established a correlation between pruning, representation alignment, and OOD accuracy. To test whether correcting geometric distortions explains the performance gains, we insert an inverse surrogate $W_\ell^{-1}$ after a layer selected for pruning. This map approximately sends the distorted post-layer representation back toward the representation that would have been obtained if the layer had been removed. If this recovers the pruned model's performance, then undoing the layer's geometric distortion is sufficient to recover much of the OOD gain.

We fit matrices $W^D_\ell$ for layers removed from our Llama 8B math expert, using \method{} with OOD MMLU as the target distribution. Figure~\ref{fig:layer23-interventions} shows the result of inserting $W^{-1}_{24}$ after layer~24 and compares it with alternative matrix interventions. The full model scores $36.0\%$; removing layer~24 raises accuracy to $37.8\%$; inserting $W^{-1}_{24}$ yields $37.5\%$, nearly matching the pruned model. MMLU improvement (+7.4) is obtained by pruning multiple layers selected by TALE.  Extending the inverse-surrogate intervention to the complete sequence of layers\footnote{ The sequence is $(24, 22, 25, 21, 10)$ using 1-based indexing.}, we first inserted the inverse surrogate for layer 24 in the full model, then removed layer 24 and inserted the inverse surrogate for layer 22, and continued this process sequentially for the remaining selected layers.

\begin{table}[t]
    \centering
    \small
    \setlength{\tabcolsep}{5pt}
    \begin{tabular}{@{}crr@{}}
        \toprule
        \shortstack{TALE pruning order\\(1-based layer index)}
        & \shortstack{Incremental gain\\(pp)}
        & \shortstack{Cumulative gain\\(pp)} \\
        \midrule
        Layer 24 & $+1.8$ & $+1.8$ \\
        Layer 22 & $+0.7$ & $+2.5$ \\
        Layer 25 & $+0.5$ & $+3.0$ \\
        Layer 21 & $+0.2$ & $+3.2$ \\
        Layer 10 & $-0.1$ & $+3.1$ \\
        \bottomrule
    \end{tabular}
    \caption{Incremental and cumulative accuracy gains from
    the inverse-surrogate, following the TALE pruning order.
    Layer indices are 1-based.}
    \label{tab:inverse_surrogate_gains}
\end{table}

The results in Table \ref{tab:inverse_surrogate_gains} show that pruned model OOD accuracy can be largely restored by approximately undoing the geometric distortion induced by selected layers.  This causally links geometric information to pruning's effect on model performance. Further rescaling experiments in Appendix~\ref{sec:alpha-sweep} show that topology changes, residual-stream interactions, and sample-level redundancy do not by themselves explain the observations.

\subsection{Summary}
\label{sec:analysis_summary}

Our results support a geometric mechanism for \method{}'s OOD gains. Distribution shift changes the model's layerwise representation geometry, producing profiles that deviate from those induced by the adapted distribution. Some layers amplify this mismatch and distort the prediction trajectory. \method{} improves OOD performance by attenuating or removing such layers, thereby aligning OOD representations with the geometry on which the model performs well.

Geometry also explains the ID/OOD asymmetry observed in Section~\ref{sec:ood}. On ID data, the full model's layers are calibrated to the task distribution, so pruning removes useful transformations, degrades performance and distorts ID geometric statistics. On OOD data, the same transformations can become miscalibrated amplifiers. In that regime, pruning can improve accuracy by correcting the representation trajectory rather than by increasing model capacity or retraining the parameters.

\section{Conclusion}
\label{sec:conclusion}
We have given a novel explanation for why inference-based pruning methods work, and when and where they yield gains.  Concentrating on the layer pruning method TALE, we have shown that it substantially improves performance on OOD data, but not on ID data.  We have also provided a geometric explanation based on empirical geometric statistics, and we have shown a causal connection between model performance and those statistics.  Finally, we have shown that layer pruning can adjust the local geometry induced by OOD data to align with the adapted task geometry.  

\begin{ack}
We acknowledge funding from the ANITI IA Cluster (ANR-23-IACL-0002), the SARER and Summ-RE projects (ANR-20-CE23-0017), and the COCOBOTS project (ANR-21-FAI2-0005). We thank CALMIP for providing computational resources under grant 2016-P23060.
\end{ack}

\bibliographystyle{plainnat}

\bibliography{tale-bis,custom}

\appendix

\newpage

\section{Limitations}

Our results support a geometric explanation for why task-aware layer pruning can improve performance under distribution shift. Several limitations, however, remain.

First, the distinction between in-distribution and out-of-distribution evaluation is most precise in our controlled regression experiments, where the training and test distributions are explicitly specified. For pretrained and fine-tuned language models, the pretraining distribution is not directly observable. We therefore operationalize the ID/OOD distinction through task specialization, treating the fine-tuning task as the adapted distribution and evaluating other benchmarks as shifted inputs. This captures an important practical form of distribution shift, but clearly does not exhaust all notions of OOD generalization.

Second, our geometric analysis relies on hidden-state norms, pairwise token distances, variance structure, and local linear surrogates. These quantities are interpretable and allow comparisons across layers, datasets, and pruning interventions, but they do not fully characterize the representation distribution. The Hessian characterization does better but it is also just a partial view of the geometry.  In particular, it is possible given our analysis that two representation distributions agree on these summaries while differing along task-relevant directions.

Third, our causal interventions are necessarily approximate. The inverse-surrogate experiments show that undoing the local geometric effect of selected layers can recover much of the pruning gain, but these surrogate maps are fitted post hoc and approximate nonlinear transformer blocks only on the evaluated data distribution. Thus, these experiments provide evidence for the role of geometric distortion, but they do not constitute a complete mechanistic decomposition of the removed layers.

Finally, our large-model experiments cover two model scales and two task families, using fine-tuned math and code specialists evaluated on several shifted benchmarks. Broader validation across architectures, pretraining recipes, instruction-tuning methods, multilingual settings, long-context tasks, and open-ended generation would further test the generality of the proposed mechanism.  Further research should also broaden the study to other pruning methods.

\section{A Geometric Interpretation of Representation Alignment}\label{appendix:math}
\label{app:geometric-interpretation}

We assume that  the input task is a probability distribution over $\mathcal X$ and for a transformer with $L$ layers of fixed dimension $d$, let
\[
h_1:\mathcal X \to \mathbb R^{d}
\]
and let 
\[
h_{\ell +1}: h_1 \circ ... \circ h_\ell(\mathcal{X}) \to \mathbb R^{d}
\]
be the representation map that sends an input $x$ to its hidden state at layer
$\ell$. 

\hidden{}

Each input distribution $P$ over $\mathcal X$ induces a layerwise representation distribution
through the pushforward measure
\[
\mu_{\ell,P} = (h_\ell)_\# P.
\]
The distribution $\mu_{\ell,P}$ describes where inputs from $P$ lie in the
model's representation space at layer $\ell$.

For an
adapted task distribution $P_{\mathrm{ID}}$, the model induces a collection of
layerwise representation distributions
\[
\mathcal{H}_{\ell}^{\mathrm{ID}}
=
\{h_\ell(x):x\sim P_{\mathrm{ID}}\}.
\]
These distributions define
hidden-state norms, pairwise token distances, and variance profiles, which in turn define the model's \emph{adapted representation
geometry} for the task at each level.

We note that this region is task specific.  An input drawn from a distribution appropriate for a task $T_j$ distinct from $T_i$ with its own ID distribution will typically define an adapted representation geometry for $T_j$ that is different from that for $T_i$.  Henceforth, we fix a particular adapted task $T_i$.

An out-of-distribution input distribution (for $T_i$) $P_{\mathrm{OOD}}$, delivers a similar but distinct layerwise representation distribution:

\[
\mathcal{H}_{\ell}^{\mathrm{OOD}}
=
\{h_\ell(x):x\sim P_{\mathrm{OOD}}\}.
\]
Pruning or attenuating such a layer replaces the update with a reduced 
contribution:
\[
h_{\ell+1}
=
h_\ell+\alpha\Delta_\ell^{\mathrm{res}}(h_\ell),
\qquad
0\leq \alpha \leq 1,
\]
Layer deletion occurs when $\alpha=0$. 

We have established empirically that $\mathcal{H}_{\ell}^{\mathrm{OOD}}$ typically differs from
$\mathcal{H}_{\ell}^{\mathrm{ID}}$, that this difference can lead to degradation of model accuracy on OOD data, and that pruning improves accuracy when
it reduces this discrepancy.
 
This difference manifests itself via inflated norms,
pairwise distances, as well as differences in variance structure or in classification margins.  Note that the important quantity is not,e.g., norm size itself but
mismatch relative to the adapted representation geometry.

\subsection{OOD mismatch as distance from adapted geometry}

We now define this mismatch more precisely.
OOD mismatch is defined by comparing the OOD representation distribution
$\mu_{\ell,\mathrm{OOD}}$ with the adapted ID representation distribution
$\mu_{\ell,\mathrm{ID}}$.  One can compare distributions directly using a probability metric
$\mathcal D$, such as Wasserstein distance:
\hidden{}
Alternatively, one can compare finite-dimensional summaries of those distributions.
Let
\[
S_\ell(P)
=
\Big(
m_\ell(P),\,
r_\ell(P),\,
v_\ell(P),\,
C_\ell(P)
\Big)
\]
denote a tuple of representation statistics at layer $\ell$, where
$m_\ell(P)$ may be a mean norm, $r_\ell(P)$ a median pairwise token distance,
$v_\ell(P)$ a variance or spread statistic, and $C_\ell(P)$ a covariance or
spectral summary. %

  \begin{Definition}
A layerwise discrepancy measure over an ID and OOD distribution is the function
\[
D_\ell(P_{\mathrm{OOD}},P_{\mathrm{ID}})
=
d_S\!\left(
S_\ell(P_{\mathrm{OOD}}),
S_\ell(P_{\mathrm{ID}})
\right),
\]
where $S_\ell(P)$ denotes a tuple of representation statistics at layer $\ell$, and $d_S$ is a suitable
distance over these tuples.  
\end{Definition}

 The experiments in the paper primarily estimate the 
discrepancy over representation statistics. In the controlled regression setting and in the LLM setting,
OOD inputs produce geometric profiles that deviate from the ID profile.
The OOD representation statistics become mismatched relative to the adapted
geometry. Depending on the direction of distribution shift, alignment may
require contraction or expansion.

We now turn to layer analysis.  As in the body of the paper, we take a the semantic effect of a transformer layer $\ell_{n+1}$ to be a residual map:
\begin{Definition}
\[
h_{\ell_{n+1}}
=
h_{\ell_n}+\Delta_{\ell_{n+1}}^{\mathrm{res}}(h_{\ell_n}),
\]
where $\Delta_{\ell_{n+1}}^{\mathrm{res}}$ is the residual update produced by the
attention and MLP blocks of $\ell_{n+1}$. 
\end{Definition}

The same layer map $h_{\ell_{n+1}}$ can behave differently on different regions of
representation space. On representations drawn from
$\mathrm{ID}$, the residual update typically calibrates and refines the output representation: it moves hidden states in directions useful for the adapted task.
On representations drawn from $\mu_{\ell,\mathrm{OOD}}$ if this layer is a candidate for elimination from the pruning method, the same residual
update amplifies the mismatch: it moves hidden states farther away
from the adapted geometry.  Formally, layer $\ell$ increases the discrepancy between OOD and ID representation
statistics with respect to the discrepancy
measure $\mathcal D$ when:
\[
d_S\!\left(
S_{\ell+1}(P_{\mathrm{OOD}}),
S_{\ell+1}(P_{\mathrm{ID}})
\right)
>
d\!\left(
S_{\ell}(P_{\mathrm{OOD}}),
S_{\ell}(P_{\mathrm{ID}})
\right),
\]

\paragraph{Local linearization and directional amplification}

To understand how a layer can amplify mismatch, consider a local linearization
of the residual map around the representations induced by a distribution $P$.
\hidden{}
We fit the best linear surrogate on data from $P$:
\[
A_{\ell,P}
=
\arg\min_A
\sum_i
\|A z_i - h_\ell(z_i)\|_2^2,
\qquad
z_i\sim \mu_{\ell,P}.
\]
The fitted map $A_{\ell,P}$ is not assumed to describe the layer globally. It
only describes the layer's average local action on the representations produced
by distribution $P$.

This local view explains why harmfulness is distribution-dependent. The same
layer weights can yield different effective linear maps on ID and OOD
representations:
\[
A_{\ell,\mathrm{ID}}
\neq
A_{\ell,\mathrm{OOD}}.
\]
If the OOD representations lie in directions where the layer has large gain,
then the layer can expand those directions even if it behaves benignly on ID
representations. A simple diagnostic is the on-data gain:
\[
g_{\ell,P}(z)
=
\frac{\|h_\ell(z)\|_2}{\|z\|_2},
\]
or, for the fitted linear surrogate,
\[
\tilde g_{\ell,P}(z)
=
\frac{\|A_{\ell,P}z\|_2}{\|z\|_2}.
\]
A large gain on OOD data, especially when concentrated in a small number of
directions, indicates that the layer acts as a low-rank amplifier for that
distribution. We observed this behavior in the linear-surrogate analysis:
the same layer can have moderate, diffuse gain on near-ID inputs and large,
concentrated gain on far-OOD inputs.
\subsection{Variance structure}
We examined two variants: a \emph{trace-normalized} version, which divides each
covariance by its trace and therefore isolates differences in \emph{shape and
orientation}; and an \emph{unnormalized} version, which additionally reflects
differences in overall \emph{scale}.

Using relative covariance to analyze distortions in the variance structure of the models prior to and post pruning reveals an important effect of pruning on variance structure.  
\begin{table}[!ht]
\footnotesize
\setlength{\tabcolsep}{3pt}
\centering
\begin{tabular}{rcccc}
\toprule
Layer & ID vs ID (F) & ID vs OOD (F) & ID vs ID (P) & ID vs OOD (P) \\
\midrule
0 & $(2.00 \pm 0.00)\times 10^{-4}$ & $(2.00 \pm 0.00)\times 10^{-4}$ & $(2.00 \pm 0.00)\times 10^{-4}$ & $(2.00 \pm 0.00)\times 10^{-4}$ \\
1 & $(2.00 \pm 0.50)\times 10^{-3}$ & $(1.46 \pm 0.003)\times 10^{-1}$ & $(2.00 \pm 0.00)\times 10^{-4}$ & $(2.00 \pm 0.00)\times 10^{-4}$ \\
2 & $(6.50 \pm 0.50)\times 10^{-3}$ & $(2.17 \pm 0.005)\times 10^{-1}$ & $(1.11 \pm 0.05)\times 10^{-2}$ & $(2.22 \pm 0.009)\times 10^{-1}$ \\
3 & $(1.45 \pm 0.11)\times 10^{-2}$ & $(3.12 \pm 0.008)\times 10^{-1}$ & $(2.67 \pm 0.06)\times 10^{-2}$ & $(3.90 \pm 0.019)\times 10^{-1}$ \\
4 & $(2.85 \pm 0.13)\times 10^{-2}$ & $(4.28 \pm 0.008)\times 10^{-1}$ & $(4.41 \pm 0.08)\times 10^{-2}$ & $(4.51 \pm 0.014)\times 10^{-1}$ \\
5 & $(4.08 \pm 0.21)\times 10^{-2}$ & $(\mathbf{4.38} \pm 0.02)\times 10^{-1}$ & $(4.81 \pm 0.12)\times 10^{-2}$ & $(\mathbf{4.36} \pm 0.01)\times 10^{-1}$ \\
6 & $(6.06 \pm 0.20)\times 10^{-2}$ & $(\mathbf{5.12} \pm 0.023)\times 10^{-1}$ & $(6.07 \pm 0.14)\times 10^{-2}$ & $(\mathbf{4.47} \pm 0.021)\times 10^{-1}$ \\
7 & $(7.69 \pm 0.13)\times 10^{-2}$ & $(\mathbf{5.61} \pm 0.028)\times 10^{-1}$ & $(7.14 \pm 0.13)\times 10^{-2}$ & $(\mathbf{4.97} \pm 0.038)\times 10^{-1}$ \\
8 & $(9.80 \pm 0.16)\times 10^{-2}$ & $(\mathbf{6.01} \pm 0.035)\times 10^{-1}$ & $(8.45 \pm 0.10)\times 10^{-2}$ & $(\mathbf{5.00} \pm 0.029)\times 10^{-1}$ \\
9 & $(1.15 \pm 0.018)\times 10^{-1}$ & $(\mathbf{6.32} \pm 0.034)\times 10^{-1}$ & $(1.02 \pm 0.013)\times 10^{-1}$ & $(\mathbf{5.40} \pm 0.027)\times 10^{-1}$ \\
10 & $(1.32 \pm 0.02)\times 10^{-1}$ & $(\mathbf{6.61} \pm 0.031)\times 10^{-1}$ & $(1.15 \pm 0.015)\times 10^{-1}$ & $(\mathbf{5.58} \pm 0.027)\times 10^{-1}$ \\
11 & $(1.45 \pm 0.023)\times 10^{-1}$ & $(\mathbf{6.85} \pm 0.035)\times 10^{-1}$ & $(1.25 \pm 0.017)\times 10^{-1}$ & $(\mathbf{5.60} \pm 0.027)\times 10^{-1}$ \\
12 & $(1.32 \pm 0.027)\times 10^{-1}$ & $(\mathbf{6.42} \pm 0.029)\times 10^{-1}$ & $(1.39 \pm 0.018)\times 10^{-1}$ & $(\mathbf{5.20} \pm 0.024)\times 10^{-1}$ \\
\bottomrule
\end{tabular}
\caption{Results for relative covariance distance on normalized ID vs. ID and ID vs. OOD on Full (F) and Pruned (P) models on regression tasks.  how the pruned model lowers covariance distance on ID vs. OOD and turns it toward the lower ID vs. ID covariance.}
\label{tab:rel-covariance-regression}
\end{table}

Table \ref{tab:rel-bijection} shows an exact correlation between the MSE of a model with that layer pruned and the covariance measure.  The scaling argument in Section \ref{sec:alpha-sweep} further supports that the decrease in OOD accuracy that pruning corrects comes from the
magnitude and direction of the residual update, which is what we measure with relative covariance, rather than from a
discrete architectural operation.

\begin{table}[!ht]
\small
\centering
\begin{tabular}{rrrl}
\toprule
Layer & MSE (layer pruned) & Ratio & Lowers variance distortion? \\
\midrule
 1 & $2.92 \times 10^{-1}$ & $0.57\times$ & yes \\
 2 & $6.22 \times 10^{-1}$ & $1.21\times$ & no  \\
 3 & $5.16 \times 10^{-1}$ & $1.00\times$ & no  \\
 4 & $4.61 \times 10^{-1}$ & $0.90\times$ & yes \\
 5 & $5.42 \times 10^{-1}$ & $1.06\times$ & no  \\
 6 & $4.99 \times 10^{-1}$ & $0.97\times$ & yes \\
 7 & $4.98 \times 10^{-1}$ & $0.97\times$ & yes \\
 8 & $4.76 \times 10^{-1}$ & $0.93\times$ & yes \\
 9 & $4.97 \times 10^{-1}$ & $0.97\times$ & yes \\
10 & $5.04 \times 10^{-1}$ & $0.98\times$ & yes \\
11 & $4.76 \times 10^{-1}$ & $0.93\times$ & yes \\
12 & $4.89 \times 10^{-1}$ & $0.95\times$ & yes \\
\bottomrule
\end{tabular}
\caption{Results showing a bidirectional correspondence between
lowering MSE and lowering variance structure distortion.
Ratios are rounded to two decimal places; the yes/no labels
reflect the unrounded values.}
\label{tab:rel-bijection}
\end{table}
For the regression case, we can fully express the difference between the MSE of pruned and full models in terms of covariance and the decoding, showing a mathematical connection between this geometric statistic and model error rates.

\begin{Proposition}
\[
\begin{aligned}
\mathrm{MSE}_P - \mathrm{MSE}_F
  &= w^\top (\Sigma^P - \Sigma^F) w \\
  &\quad - 2w^\top \bigl(
      \operatorname{Cov}(h_L^P, y) - \operatorname{Cov}(h_L^F, y)
    \bigr) \\
  &\quad + \bigl(w^\top \mathbb{E}[h_L^P] + b - \mathbb{E}[y]\bigr)^2 \\
  &\quad - \bigl(w^\top \mathbb{E}[h_L^F] + b - \mathbb{E}[y]\bigr)^2 .
\end{aligned}
\]
\end{Proposition}

\hidden{}

Recall that \[
\mathrm{MSE}=\mathbb{E}[(\hat y-y)^2]
\]
and 
\[
\mathbb E[Z^2]
=
\operatorname{Var}(Z)+(\mathbb E[Z])^2.
\]

Now let $Z=\hat y-y$; then 
 \[
\mathrm{MSE}
=
\operatorname{Var}(\hat y-y)
+
\left(\mathbb E[\hat y-y]\right)^2.
\]

We can use 
\[
\operatorname{Var}(X-Y)
=
\operatorname{Var}(X)
+
\operatorname{Var}(Y)
-
2\operatorname{Cov}(X,Y).
\]
and substituting  we get
\[
\operatorname{Var}(\hat y-y)
=
\operatorname{Var}(\hat y)
+
\operatorname{Var}(y)
-
2\operatorname{Cov}(\hat y,y).
\]
So
\begin{equation}\label{mse-basic} MSE = Var(\hat{y}) + Var(y) -2Cov(\hat{y},y) + (\mathbb{E}[\hat{y}] - \mathbb{E}[y])^2
\end{equation} 

The link between hidden state variance structure and the output depends also on the regression head.  In our case it is linear so the trained model for input computes 
\begin{equation} \label{linear-decoder} \hat{y} = w^\top h_L +b
\end{equation}
where $h_L$ is the final output of the last layer.  Since no weights are recomputed in a pruned model, the pruned and full model estimates
\[ \hat{y}^f = w^\top h^F_L +b \ \ \mathit{and} \ \ \hat{y}^P = w^\top h^p_L +b
\]
 use the same parameters.  
 Now since b is a constant,
 \[
\operatorname{Var}(\hat y)
=
\operatorname{Var}(w^\top h).
\]
Let $\Sigma_h
=
\operatorname{Cov}(h)
=
\mathbb E[
(h-\mathbb E[h])
(h-\mathbb E[h])^\top
]$. Then 
\[Var(w^\top h) = w^\top \Sigma_h w \]
which gives 
\begin{equation} \label{var1} Var(\hat{y}) = w^\top \Sigma_h w
\end{equation}

 Now considering the covariance term, 
 \[
\operatorname{Cov}(\hat y,y)
=
\operatorname{Cov}(w^\top h+b,y).
\]
Again we can ignore b, and so:\
\[
\operatorname{Cov}(\hat y,y)
=
\operatorname{Cov}(w^\top h,y).
\]
Since covariance is linear,
\begin{equation} \label{covar}
\operatorname{Cov}(w^\top h,y)
=
w^\top\operatorname{Cov}(h,y).
\end{equation}

Now coming back to the expectation term, we have: 
\[
\mathbb E[\hat y]
=
\mathbb E[w^\top h+b]
=
w^\top\mathbb E[h]+b.
\]
and so 
\[
\mathbb{E}[\hat y]-\mathbb{E}[y]
=
w^\top\mathbb E[h]+b-\mathbb E[y].
\]
Now substituting the expectation term result, \ref{var1} and \ref{covar} into \ref{mse-basic} we get:
\[MSE = w^T\Sigma_hw + Var(y) - 2w^\top Cov(h,y) + (w^\top\mathbb{E}[h_L] + b - \mathbb{E}[y])^2\]

 \hidden{} 
 
 We can plug the two final hidden states $h^F$ and $h^P$ for the full and pruned models into this equation Because we compare pruned and full models on the same OOD test samples, we get unchanged w, b and a complete description of MSE shift in terms of covariance, assuming the linear read out:

\[
\begin{aligned}
\mathrm{MSE}_P - \mathrm{MSE}_F
  &= w^\top \bigl(\Sigma_{h_L}^P - \Sigma_{h_L}^F\bigr) w \\
  &\quad - 2w^\top \bigl(
      \operatorname{Cov}(h_L^P, y) - \operatorname{Cov}(h_L^F, y)
    \bigr) \\
  &\quad + \bigl(w^\top \mathbb{E}[h_L^P] + b - \mathbb{E}[y]\bigr)^2 \\
  &\quad - \bigl(w^\top \mathbb{E}[h_L^F] + b - \mathbb{E}[y]\bigr)^2 .
\end{aligned}
\]

 \hidden{}

\hidden{}
\hidden{}

\hidden{}

\hidden{}

\section{TALE Improves OOD Performance Under Both Adam and Muon}
\label{appendix:alternatives}

Could the OOD gains observed with \method{} be specific to Adam's coordinate-wise adaptive updates? To test this possibility, we train the same polynomial regression architecture with Muon \citep{jordan2024muon}. We then apply \method{} and compare the validation loss of the pruned model with that of the full Muon-trained model at each shift scale $\sigma$.

At $\sigma=1$, \method{} retains all layers in both models. At every tested scale from $\sigma=3$ to $\sigma=10$, pruning reduces validation loss relative to the corresponding full model under both optimizers (Tables~\ref{muon_coeff}). For the Muon-trained model, \method{} removes the same four layers $\{6,8,9,12\}$ throughout this range, reducing loss by $17.7\%$ at $\sigma=3$ and $8.8\%$ at $\sigma=10$.

\hidden{}
\hidden{}

\hidden{}
\hidden{}

\hidden{}

\begin{table}[ht!]
\footnotesize
\setlength{\tabcolsep}{3pt}
\centering
\begin{tabular}{ccccl}
\toprule
$\sigma$ & \# pruned (best) & Val Best & Best/Full & Best pruned layers \\
\midrule
1 & 0 & $1.74\times 10^{-4}$ & $1.00\times$ & none \\
2 & 4 & $0.12$ & $1.02\times$ & $[6, 8, 9, 12]$ \\
3 & 4 & $1.09$ & $0.82\times$ & $[6, 8, 9, 12]$ \\
4 & 4 & $2.41$ & $0.84\times$ & $[6, 8, 9, 12]$ \\
5 & 4 & $5.82$ & $0.84\times$ & $[6, 8, 9, 12]$ \\
6 & 4 & $7.35$ & $0.85\times$ & $[6, 8, 9, 12]$ \\
7 & 4 & $12.24$ & $0.87\times$ & $[6, 8, 9, 12]$ \\
8 & 4 & $16.74$ & $0.90\times$ & $[6, 8, 9, 12]$ \\
9 & 4 & $24.97$ & $0.90\times$ & $[6, 8, 9, 12]$ \\
10 & 4 & $34.92$ & $0.91\times$ & $[6, 8, 9, 12]$ \\
\bottomrule
\end{tabular}
\caption{Best Muon models with samples $x \in \mathcal{U}(-1,1)$ and coefficients in $\mathcal{U}(-\sigma,\sigma)$ show no pruning at $\sigma = 1$, followed by stable multi-layer pruning beginning at $\sigma = 2$ and converging toward recurrent dropped sets such as $[6,10,11,12]$, with best/full ratios improving toward $0.9299$. As distribution shift increases, pruning becomes progressively beneficial even under Muon, indicating that pruning targets OOD-induced representational distortion rather than optimizer-specific redundancy.}
\label{muon_coeff}
\end{table}

\hidden{}
\hidden{}

\newpage

\hidden{}

\section{Training Setup for Regression experiments}
\label{sec:expconfig}

\subsection{Architecture and Token Representation}

All models in Section~\ref{sec:ood-regression} use the same decoder-only Transformer architecture as in~\cite{naim2025analyzing,naim2025re}, with 12 layers, 8 attention heads, and a hidden dimension of $d=256$. We train all models from scratch and omit dropout since prompts are resampled at every step. Scalar tokens are projected into $\mathbb{R}^{256}$ via a learned encoder $W_{\mathrm{enc}}$, processed autoregressively, and mapped back to scalar predictions through a learned readout $W_{\mathrm{dec}}$.

\subsection{Training}

\paragraph{Hyperparameters.} Table~\ref{tab:hyperparams} summarizes the main training configuration.

\begin{table}[h]
\centering
\caption{Training hyperparameters.}
\label{tab:hyperparams}
\begin{tabular}{lc}
\toprule
\textbf{Hyperparameter} & \textbf{Value} \\
\midrule
Optimizer       & Adam            \\
Learning rate   & $10^{-4}$       \\
Batch size      & 64              \\
Training steps  & 500k            \\
Max context length $k$ & 40       \\
\bottomrule
\end{tabular}
\end{table}

\paragraph{Prompt sampling.} At each step, a prompt is constructed by 
(i) sampling $g \sim \mathcal{D}_{\mathcal{F}}$, 
(ii) drawing inputs $x_1,\dots,x_{k+1} \sim \mathcal{D}_{\mathcal{I}}$ i.i.d., and 
(iii) evaluating $g$ to form the sequence. 
The training loss is:
\begin{equation}
    \mathcal{L} = \frac{1}{k}\sum_{i=1}^{k}\bigl(\hat{y}_i - g(x_i)\bigr)^2.
\end{equation}

\paragraph{Curriculum.} Context length starts at 11 examples and grows by 2 every 2{,}000 steps. For multi-degree experiments, polynomial classes are additionally introduced in order of increasing degree.

\subsection{Evaluation}

We evaluate ICL performance using five random seeds (42, 123, 456, 789, and 1011). For each seed, we sample 100 functions from the test distribution and, for each function, 64 sequences of 41 inputs drawn independently from the test input distribution. At each position, the model predicts the output for the current input given the preceding input-output pairs. We score predictions only once the context contains enough examples to identify the polynomial, which requires one more example than its degree. We report the mean squared error averaged over scored positions, sequences, functions, and seeds. All models are evaluated on the same functions and inputs for each seed.

\section{Per-Benchmark $L_2$ Norm Computation and Calibration}
\label{sec:l2-implementation}

To make the $L_2$ analysis comparable across benchmarks of differing input length, vocabulary, and prompt structure, we adopt a uniform extraction-and-aggregation pipeline that is calibrated independently for each benchmark before any pruning intervention is applied. This subsection details the procedure.

\paragraph{Calibration set.}
For each benchmark $\mathcal{T}$, we draw a fixed calibration set
$\mathcal{C}_{\mathcal{T}} = \{x^{(1)}, \dots, x^{(N)}\}$ of $N{=}50$ raw question texts from the canonical evaluation split: the \texttt{test} split for MMLU and BoolQ~\cite{boolq}. For BoolQ, we concatenate passage and question into a single string. The samples are taken in dataset order rather than randomly so that the calibration set is reproducible and identical across pruning iterations and across the baseline measurement. Each text is tokenized with the model's native tokenizer, truncated to a maximum of $512$ tokens, and fed through the model in evaluation mode (\texttt{torch.no\_grad}, \texttt{fp16}) with \texttt{output\_hidden\_states=True} so that the post-block residual stream at every transformer layer is exposed.

\paragraph{Last-token representation.}
Let $h^{(\ell)}(x) \in \mathbb{R}^{d}$ denote the hidden state of the \emph{last} input token of $x$ at the output of layer $\ell$, with $d$ the model's hidden dimension. We focus on the final layer $\ell{=}L$, since this is the representation immediately consumed by the language modeling head and therefore the locus where any disruption introduced by removing an intermediate layer must ultimately manifest. We use the last-token position because in causal decoder-only models it is the only position whose hidden state attends to the entire input.

\paragraph{Two scalar metrics.}
Given the calibration set $\mathcal{C}_{\mathcal{T}}$, we instantiate two scalar summaries of the geometry of $\{h^{(L)}(x^{(i)})\}_{i=1}^{N}$.

The first is the mean last-token norm,
\begin{equation}
    \mathrm{Norm}(\mathcal{C}_{\mathcal{T}})
    \;=\;
    \frac{1}{N} \sum_{i=1}^{N} \bigl\| h^{(L)}(x^{(i)}) \bigr\|_{2},
    \label{eq:l2-norm-single}
\end{equation}
which captures the typical activation magnitude at the readout position.

The second, which we use as our default, is the mean pairwise $L_2$ distance between calibration examples,
\begin{equation}
    \mathrm{PD}(\mathcal{C}_{\mathcal{T}})
    \;=\;
    \frac{2}{N(N-1)} \sum_{1 \leq i < j \leq N}
    \bigl\| h^{(L)}(x^{(i)}) - h^{(L)}(x^{(j)}) \bigr\|_{2}.
    \label{eq:l2-pairwise}
\end{equation}
We prefer the pairwise distance because it measures how spread out the model's representations of distinct inputs are, rather than how large any single vector is. A drop in $\mathrm{PD}$ after pruning therefore signals that the pruned layers were contributing to the \emph{separation} of inputs at the readout, which is the property we link to OOD behavior in Section~\ref{sec:analysis}; the single-vector norm in \eqref{eq:l2-norm-single} can in contrast move purely because of changes to a global scale factor (e.g.\ residual-stream growth) without any change in discriminability. The summation in \eqref{eq:l2-pairwise} is restricted to the strict upper triangle to avoid double counting and the $i{=}j$ diagonal.
\section{Training and Fine tuning experimental setups for LLMs}
\label{appendix:training-llms}

\subsection{GPT OSS 120B}
We fine-tune gpt-oss-120b using parameter-efficient LoRA adapters~\cite{hu2021loralowrankadaptationlarge} applied to both the attention projections (\texttt{q\_proj}, \texttt{k\_proj}, \texttt{v\_proj}, \texttt{o\_proj}) and the MoE expert projections (\texttt{gate\_up\_proj}, \texttt{down\_proj}). We use rank $r{=}16$ with $\alpha{=}32$ and dropout $0.05$, yielding a small fraction of trainable parameters relative to the 120B base model.

Training is performed in bfloat16 with FSDP full sharding~\cite{zhao2023pytorchfsdpexperiencesscaling} across 4 nodes $\times$ 2 A100-80GB GPUs (8 GPUs total), wrapping the model at the \texttt{GptOssDecoderLayer} granularity. Inputs are formatted with the model's native Harmony chat template via \texttt{apply\_chat\_template}, with a task-specific system prompt and a maximum sequence length of 2048 tokens. We train for one epoch on a 10{,}000-example subset with per-device batch size 1 and gradient accumulation of 8 (effective global batch size 64), using the AdamW optimizer with a cosine learning rate schedule, peak learning rate $1\mathrm{e}{-}4$, weight decay $0.01$, gradient clipping at $1.0$, and warmup over $5\%$ of training steps. Gradient checkpointing is enabled to fit activations in memory. We use eager attention for FSDP compatibility and disable KV caching during training. The resulting LoRA adapter is retained without merging into the base MXFP4 weights, since merging would force dequantization of the affected MoE blocks.

\subsection{Llama 3.1 8B}
We fine-tune \texttt{meta-llama/Llama-3.1-8B-Instruct} on a domain-specific mathematics dataset to obtain $M_{\text{math}}$, the math-specialized model used throughout our experiments. The fine-tuning procedure is designed to specialize the model on mathematical reasoning while remaining computationally tractable on commodity hardware.

\paragraph{Training Data.}
{ We used 20{,}000 problems randomly sampled from NuminaMath-CoT~\cite{numina_math_datasets}.}%

For coding , we take code alpaca dataset that has 20k datapoints out of which we select 10k for training. Each example consists of a triple $(\text{problem}, \text{solution}, \text{answer})$, where solutions follow chain-of-thought reasoning and final answers are appended in the format \texttt{\#\#\#\# [answer]}.

\paragraph{Prompt Format.}
We use Llama 3.1's native chat template via \texttt{tokenizer.apply\_chat\_template} with the following structure:
\begin{itemize}
    \item \textbf{System prompt:} ``You are a mathematical problem-solving assistant. Solve the given math problem step by step with clear reasoning. Show all your work and calculations. At the end, provide your final answer after `\#\#\#\#' in the format: \#\#\#\# [answer].''
    \item \textbf{User content:} ``Problem: \{problem\}\textbackslash n\textbackslash nSolve this step-by-step and provide the final answer after \#\#\#\#.''
    \item \textbf{Assistant content:} The full chain-of-thought solution with the final boxed answer.
\end{itemize}

\paragraph{LoRA Configuration.}
To enable efficient fine-tuning of the 8B-parameter model, we use Low-Rank Adaptation (LoRA)~\cite{hu2021loralowrankadaptationlarge} with rank $r = 64$, $\alpha = 16$, and dropout $0.1$. LoRA adapters are applied to all linear projections (\texttt{target\_modules="all-linear"}), including attention (Q, K, V, O projections) and MLP layers (gate, up, down projections). Only the LoRA parameters are trained; the base model weights remain frozen.

\paragraph{Optimization.}
We train for 3 epochs using the paged AdamW optimizer (\texttt{paged\_adamw\_32bit}) with a peak learning rate of $2 \times 10^{-4}$, weight decay of $0.001$, and a cosine learning rate schedule with $3\%$ warmup. Gradient clipping is applied at $\|\nabla\|_2 \leq 0.3$, and bfloat16 mixed precision is used throughout training. The effective batch size is $32$ (per-device batch size of $2$, gradient accumulation of $16$). Maximum sequence length is set to $2048$ tokens.

\paragraph{Quantization for Memory Efficiency.}
In single-GPU configurations, the base model is loaded with 4-bit NF4 quantization using \texttt{bitsandbytes}, with double quantization enabled and bfloat16 compute dtype, reducing memory requirements from $\sim$16~GB (FP16) to $\sim$5~GB. In multi-node distributed configurations, we disable 4-bit quantization (incompatible with DistributedDataParallel) and instead use bf16 weights with FSDP (full sharding, automatic wrapping). Gradient checkpointing is enabled in single-GPU mode and disabled under DDP/FSDP.

\paragraph{Hardware and Distributed Setup.}
Training is performed on the CALMIP Turpan cluster, using up to 4 compute nodes each with 2 NVIDIA A100 80GB GPUs (8 GPUs total) connected via Infiniband HDR. Distributed training is launched with \texttt{torchrun} using the \texttt{c10d} rendezvous backend.

\paragraph{Adapter Merging.}
After training, the LoRA adapter is merged into the base model via \texttt{merge\_and\_unload}, producing a standalone fine-tuned model in fp16. The merged checkpoint $M_{\text{math}}$ is saved to disk and used for all downstream experiments (TALE layer dropping, alpha-rescaling, and lm-evaluation-harness evaluation).

\paragraph{Reproducibility.}
We fix random seeds ($42$) for NumPy, PyTorch, and CUDA, and set \texttt{torch.backends.cudnn.deterministic = True}. Training metrics are logged via TensorBoard with logging every $25$ steps and checkpoints saved every $500$ steps (retaining the most recent $3$ checkpoints).

\section{Table for finegrained \method performance on regression} \label{appendix:regression}
Table \ref{linear2} shows the behavior of $Ba_1$ and various $Be_\sigma$ models on more finegrained $\sigma$.  The base model continues to beat all pruning models for several finegrained distribution shifts.  But for $\sigma > 1.6$, \method delivers consistent OOD improvement.  Table \ref{linear2} shows that by shifting OOD out \method can provide us with several distinct task experts.
\begin{table}[ht!]
\footnotesize
\setlength{\tabcolsep}{3pt}
\centering
\begin{tabular}{@{}cccccccl@{}}
\toprule
$\sigma$ & ACC & both & only $Ba_1$ & only $Be_\sigma$ & neither & agg ratio & \# Pruned \\
\midrule
1 & $8.00\times 10^{-6}$ & 77 & 0 & 0 & 23 & $1.00\times$ & 0 \\
1.1 & $1.20\times 10^{-5}$ & 73 & 0 & 0 & 27 & $1.00\times$ & none \\
1.2 & $3.90\times 10^{-5}$ & 86 & 0 & 0 & 14 & $1.00\times$ & none \\
1.3 & $1.66\times 10^{-4}$ & 86 & 0 & 0 & 14 & $1.00\times$ & none \\
1.4 & $5.83\times 10^{-4}$ & 82 & 0 & 0 & 18 & $1.00\times$ & none \\
1.5 & $1.70\times 10^{-3}$ & 79 & 0 & 0 & 21 & $1.00\times$ & none \\
1.6 & $4.17\times 10^{-3}$ & 76 & 0 & 4 & 20 & $0.94\times$ & $[7]$ \\
1.7 & $8.90\times 10^{-3}$ & 74 & 0 & 2 & 24 & $0.95\times$ & $[7]$ \\
1.8 & $0.02$ & 73 & 0 & 1 & 26 & $0.95\times$ & $[7]$ \\
1.9 & $0.03$ & 72 & 1 & 0 & 27 & $0.85\times$ & $[7,10]$ \\
2 & $0.04$ & 73 & 0 & 12 & 15 & $0.85\times$ & $[7,11]$ \\
3 & $0.75$ & 68 & 0 & 3 & 29 & $0.80\times$ & $[7,10,11]$ \\
4 & $1.80$ & 71 & 0 & 2 & 27 & $0.85\times$ & $[7,11,12]$ \\
5 & $5.05$ & 65 & 0 & 4 & 31 & $0.86\times$ & $[6,10,11,12]$ \\
6 & $6.69$ & 71 & 1 & 0 & 28 & $0.88\times$ & $[6,10,11,12]$ \\
7 & $11.30$ & 68 & 0 & 6 & 26 & $0.88\times$ & $[6,10,11,12]$ \\
8 & $14.99$ & 63 & 0 & 2 & 35 & $0.90\times$ & $[6,10,11,12]$ \\
9 & $23.57$ & 59 & 0 & 6 & 35 & $0.92\times$ & $[6,10,11,12]$ \\
10 & $33.80$ & 60 & 0 & 3 & 37 & $0.93\times$ & $[6,10,11,12]$ \\
\bottomrule
\end{tabular}
\caption{TALE applied to a transformer (M1) trained with samples and coefficients from $\mathcal{U}(-1, 1)$. Test distributions $\mathcal{U}(-\sigma, \sigma)$ with $\sigma > 1$ are out-of-distribution. For each $\sigma$, 100 linear functions are classified by whether the full model ($Ba_1^+$), the TALE-pruned model ($Be_\sigma^+$), both achieve an MSE below the full model's mean-MSE threshold (ACC), whether neither does, whether $Ba_1$ does but not $Be_\sigma^+$, or whether $Be_\sigma$ but not $Ba_1$ does. The aggregate ratio (agg ratio) confirms consistent OOD improvement after pruning.}
\label{linear2}
\end{table}

\newpage

\section{Plots for NLP benchmarks: \method affects representational norm}
\label{appendix:nlp-benchmarks-norm}

As can be seen from the figures below, in each case \method where it improved performance it also lowered average pairwise L2 distance.  

We note that the norms are not the same across models or even across tasks in the same model; Llama's mean L2 distance for GSM8k is much lower than its mean L2 distance for pairs in BigBench. 

\begin{figure}[!ht] 
\includegraphics[width=7cm]{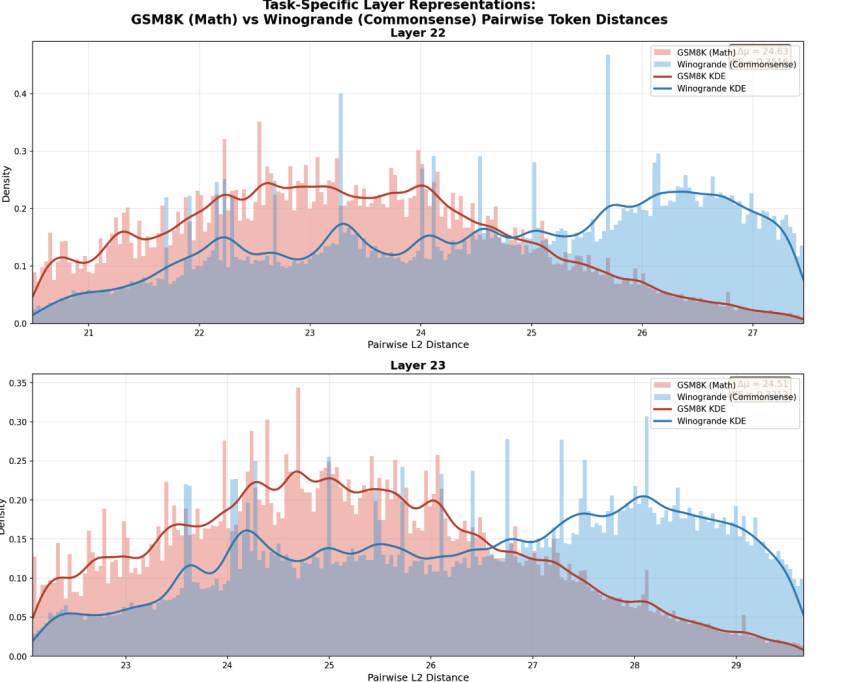}
\caption{Plot of L2 pair distances across GSM8K and Winogrande with Llama} \label{fig:comp}
\end{figure}

In general different benchmarks like GSM8k and Winogrande yield different average pairwise distances over their datasets in various layers, as illustrated for LLama 3.1 8b in Figure \ref{fig:comp}. 
The spreads between the base model and the pruned model are also quite different.  Given our other results in the paper, we take this to mean that pruning is lowering the norm towards some task adapted norm to which we do not have access.  This motivates our building and using a task adapted expert math and code models in Sections \ref{sec:math-experts} and \ref{sec:llm_geometry}.
\begin{figure}[t] \label{fig:tale-impr}
\includegraphics[width=7cm]{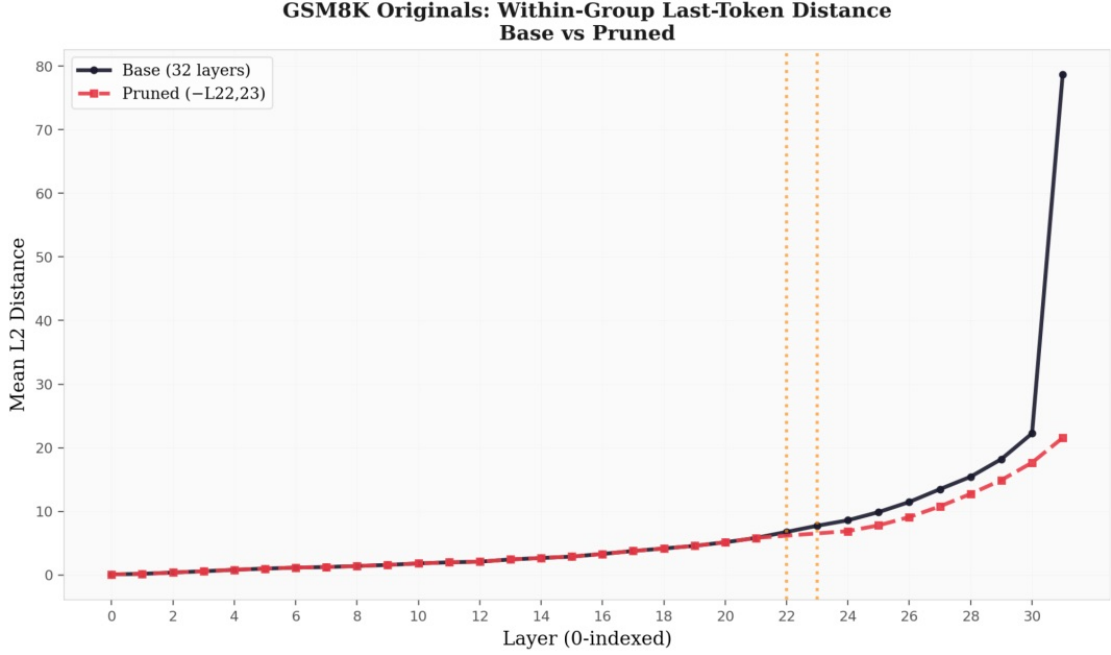}
\includegraphics[width=7cm, height=4cm]{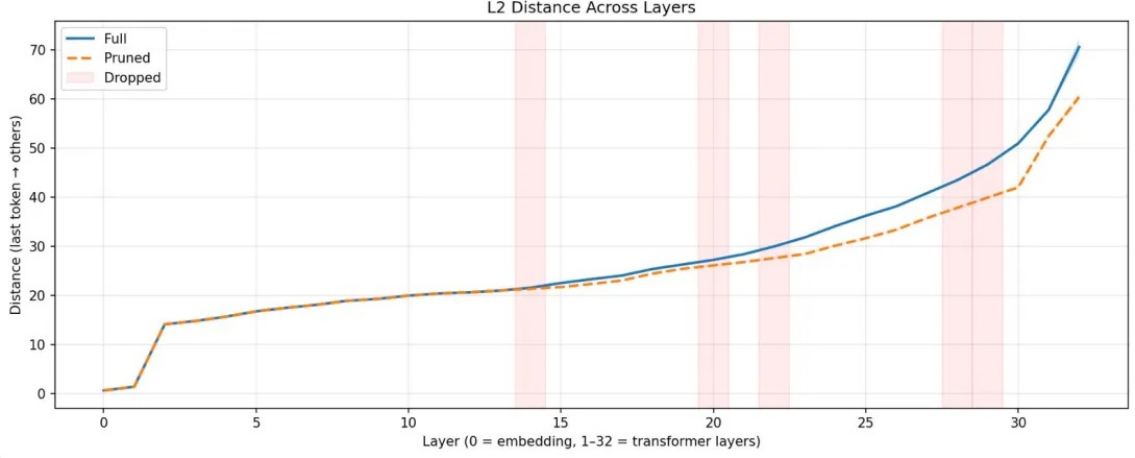}
\includegraphics[width=7cm, height=4cm]{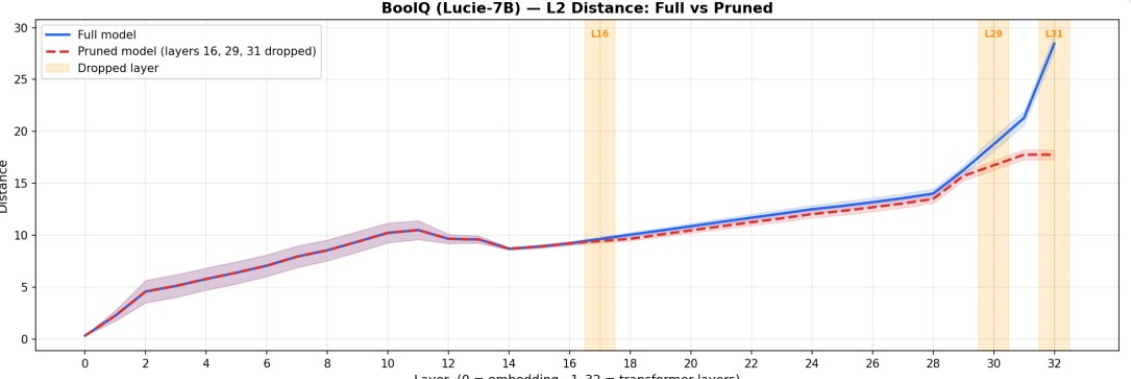}
\caption{L2 distances before and after \method pruning on GSM8k, BigBench (both on LLama 3.1 8b) and on Boolq (Lucie 7b)}
\end{figure}

\section{Linear-surrogate diagnostics: extended results}
\label{app:w-diagnostics}

This appendix gives the per-cell histograms and  layer analyses surrogate analysis in Section \ref{sec:linear_surrogate}.

\paragraph{One-sided expansion vs.\ two-sided refinement.} Section
\ref{sec:linear_surrogate} reported median norm gain only. The shape of the
gain distribution turns out to carry more information. Compare Math-L8 and
Math-L21: both undropped-vs-dropped on the same near-ID slice, both with
median gain close to~$1$ and fully diffuse update spectra (stable
ranks~$105.6$ and~$125.9$). On the median alone they appear
interchangeable. They are not. L8's on-data gain distribution is
two-sided, supported on roughly $[0.93, 1.06]$ with comparable mass below
and above~$1$: it contracts roughly as many tokens as it expands. L21's
gain distribution is one-sided, supported on $[1.00, 1.14]$ with
essentially no mass below~$1$: every token leaves the layer slightly
larger in norm than it entered.
OOD representations are
uniformly larger than ID representations at every late-depth layer.
Layers that push gain above~$1$ on every token are, by construction, the
layers that compound the gap along the residual stream. Pruning a subset
of them reverts that portion of the accumulation to identity. We can
summarise this asymmetry as a single scalar per layer,
\[
S_\ell \;=\; \frac{P(\|Wx\|/\|x\| > 1) - P(\|Wx\|/\|x\| < 1)}
                  {P(\|Wx\|/\|x\| > 1) + P(\|Wx\|/\|x\| < 1)},
\]
where $S_\ell = +1$ for strictly expanding layers, $-1$ for strictly
contracting, $0$ for balanced. On near-ID Math, Math-L21 has
$S \approx +1$ while Math-L8 has $S \approx 0$. Whether this
distinction holds across the rest of TALE's drop set
$\{10, 19, 22, 24, 25\}$ versus other undropped near-identity layers is
the natural extension of this analysis.

\begin{table}[h]
\centering
\caption{Extended diagnostics including Math-L8 and gain-distribution
spread (middle 90\%).}
\label{tab:w-diag-full}
\small
\begin{tabular}{llcrrr}
\toprule
Slice & Layer & \method drops? & Median gain & Gain spread & s-rank \\
\midrule
Religion MMLU (far-OOD)    & 3  & No  & 8.49 & $1\to 600^+$ tail & 1.02 \\
Religion MMLU (far-OOD)    & 25 & Yes & 1.09 & $0.95 \to 1.30$   & 1.11 \\
Mathematics MMLU (near-ID) & 3  & No  & 1.22 & $1.00 \to 1.40$   & 33.88 \\
Mathematics MMLU (near-ID) & 8  & No  & 0.99 & $0.93 \to 1.06$   & 105.59 \\
Mathematics MMLU (near-ID) & 21 & Yes & 1.08 & $1.00 \to 1.14$   & 125.90 \\
\bottomrule
\end{tabular}
\end{table}

\section{An additional causal argument: rescaling without pruning}
\label{sec:alpha-sweep}

Complementing our causal investigation in Section \ref{sec:surrogates-causal}, we 
consider a
continuous, non-discrete intervention that targets the layer's residual
contribution directly, without removing the layer or changing the
network's topology in any other way.

\paragraph{Method.} For a target layer~$\ell$, we replace the standard
residual update
\[
h_{\ell+1} \;=\; h_\ell +  \Delta_{\ell}(h_{\ell}))
\]
with the rescaled update
To test whether the magnitude of the residual update is causally involved, we replace the residual update at a TALE-selected layer by
\[
h_{\ell+1}
=
h_{\ell}
+
\alpha \Delta_{\ell}(h_{\ell}),
\]
where $\Delta_{\ell}$ denotes the layer's attention and MLP updates, and $\alpha \in [0,1]$. At $\alpha = 1$ the
forward pass is unchanged from baseline. At $\alpha = 0$ the layer's
contribution is fully removed---the residual stream skips the block
entirely---which is precisely what \method does when it drops layer~$\ell$.
The intermediate values trace a continuous interpolation between the two
endpoints. Crucially, no parameter is retrained, no other layer is
modified, and the network's connectivity is identical at every value of
$\alpha$. Any change in accuracy along the sweep is therefore attributable
to the magnitude of layer~$\ell$'s residual contribution alone.

\paragraph{Setup.} We run the sweep on $M_{\text{math}}$ targeting MMLU,
with $\ell$ chosen as one of the six layers \method selects when targeting
MMLU on $M_{\text{math}}$ (Section \ref{sec:math-experts}). We sweep
$\alpha \in \{0.0, 0.2, 0.4, 0.6, 0.8, 1.0\}$ and evaluate accuracy at
each point. The single-layer setup is a deliberate
underestimate of TALE's full effect: \method removes a coordinated set of
six layers found by greedy search, whose combined contribution to
accuracy is $+7.4$ points; isolating one of those six and rescaling it
alone gives a lower bound on what continuous norm-magnitude reduction
can buy.

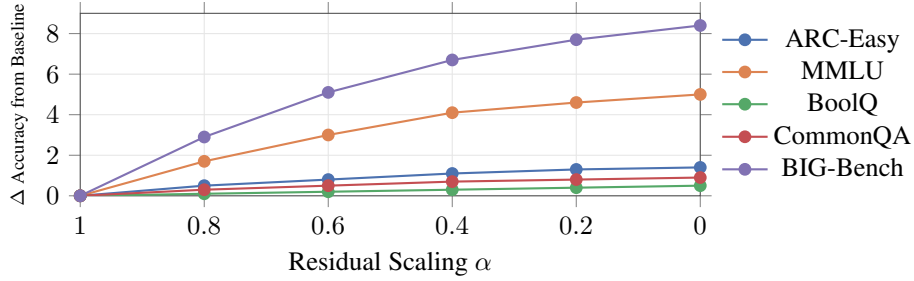
\begin{figure}[t]
\centering
\begin{tikzpicture}
\begin{axis}[
    width=0.7\linewidth,
    height=4cm,
    xlabel={Residual Scaling $\alpha$},
    ylabel={\scriptsize{$\Delta$ Accuracy from Baseline}},
    xmin=0, xmax=1,
    x dir=reverse,
    ymin=0, ymax=9,
    grid=major,
    grid style={gray!20},
    axis background/.style={fill=white},
    tick align=outside,
    axis line style={black!70},
    legend style={
        at={(1.02,0.5)},
        anchor=west,
        draw=none,
        fill=none
    },
]

\addplot[thick, color=mutedblue, mark=*] coordinates {
    (1.0,0.0) (0.8,0.5) (0.6,0.8) (0.4,1.1) (0.2,1.3) (0.0,1.4)
};
\addlegendentry{ARC-Easy}

\addplot[thick, color=mutedorange, mark=*] coordinates {
    (1.0,0.0) (0.8,1.7) (0.6,3.0) (0.4,4.1) (0.2,4.6) (0.0,5.0)
};
\addlegendentry{MMLU}

\addplot[thick, color=mutedgreen, mark=*] coordinates {
    (1.0,0.0) (0.8,0.1) (0.6,0.2) (0.4,0.3) (0.2,0.4) (0.0,0.5)
};
\addlegendentry{BoolQ}

\addplot[thick, color=mutedred, mark=*] coordinates {
    (1.0,0.0) (0.8,0.3) (0.6,0.5) (0.4,0.7) (0.2,0.8) (0.0,0.9)
};
\addlegendentry{CommonQA}

\addplot[thick, color=mutedpurple, mark=*] coordinates {
    (1.0,0.0) (0.8,2.9) (0.6,5.1) (0.4,6.7) (0.2,7.7) (0.0,8.4)
};
\addlegendentry{BIG-Bench}

\end{axis}
\end{tikzpicture}
\caption{Performance gain ($\Delta$ accuracy from baseline) as a function of residual scaling $\alpha$.}
\label{fig:residual_scaling_delta}
\end{figure}

\paragraph{Result.} Accuracy improves quasi monotonically as the layer's
residual contribution is attenuated (Table~\ref{tab:alpha-sweep},
Figure~\ref{fig:residual_scaling_delta}). Going from $\alpha = 1.0$ (baseline,
$0.367$) to $\alpha = 0.0$ (single-layer removal, $0.389$) yields a
$+2.2$ point gain on MMLU. The trajectory is essentially monotone with
just one, and it is achieved without removing the layer, without
retraining, and without modifying any other component of the network.

\begin{table}[h]
\centering
\caption{Accuracy of $M_{\text{math}}$ on MMLU as the residual contribution
of a single TALE-selected layer is rescaled by $\alpha$. The forward pass
at $\alpha = 1$ is the unmodified baseline; at $\alpha = 0$ the layer's
contribution is fully removed (equivalent to dropping the layer). All
intermediate values are continuous interpolations.}
\label{tab:alpha-sweep}
\small
\begin{tabular}{cc}
\toprule
$\alpha$ & MMLU accuracy \\
\midrule
1.00 (baseline) & 0.367 \\
0.80            & 0.374 \\
0.60            & 0.370 \\
0.40            & 0.378 \\
0.20            & 0.378 \\
0.00 (removed)  & \textbf{0.389} \\
\bottomrule
\end{tabular}
\end{table}

\paragraph{What this rules out, and what it leaves intact.} The
intervention is precise enough to discriminate among competing
explanations. \emph{First}, the gain cannot be attributed to topological
discontinuity. The forward pass at $\alpha = 0.2$ has identical
connectivity to the forward pass at $\alpha = 1.0$; only the magnitude
of one layer's contribution differs. Yet the $\alpha = 0.2$ accuracy
is already above baseline. Whatever drives the improvement is reading
the residual contribution's magnitude, not the presence or absence of
the layer as a discrete object. \emph{Second}, the gain cannot be
attributed to side-effects of layer removal interacting with surrounding
LayerNorms or attention sinks, because at intermediate $\alpha$ the
layer is still firing and its outputs are still being normalised
downstream. \emph{Third}, the gain cannot be a parameter-redundancy
effect of the kind that motivates classical pruning: redundancy
arguments predict that a layer's removal either matters or does not, not
that gradually attenuating it produces gradually larger improvements.

The intervention is consistent, on the other hand, with the our geometrical view: this
particular layer contributes a positive-on-average residual update on
OOD inputs, and reducing the magnitude of that update at test time
reduces the OOD geometric distortion the layer introduces. The
monotone $\alpha$-accuracy curve is what the magnitude-causation
hypothesis predicts and what the discrete-topology hypothesis does not.

\hidden{}
\hidden{}

\section{Norms and TALE}
\label{appendix:norms}

Complementing the figures 3 and 4, we show here the variance for these models.

The same thing hold for the x+y median distances:   
 \begin{figure}[!ht]
    \includegraphics[width=\textwidth]{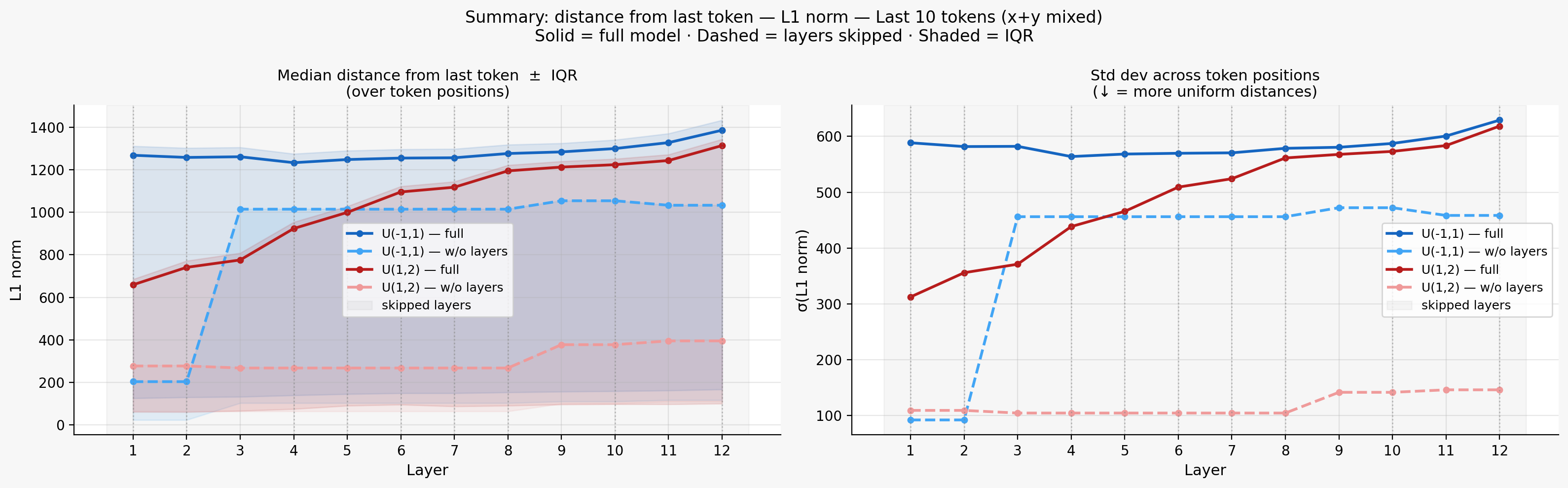}
    \caption{small transformer trained on U(1,2) with OOD data set U(-1,1). Dashed lines are performance of model \method pruned for U(-1,1) omitting layers [1, 2, 4, 5, 6, 7, 8, 10, 12].}
    \label{fig:linear-ood3}
    \end{figure} 
    The variances of the distances also show the significant shift in representational norms.  When looking at the median distances for the x entries we see much less of a difference, since the model's autoregressive "predictions" there aren't really of interest, since the inputs x are chosen randomly.  
\begin{figure}[h]
\centering
\includegraphics[width=\linewidth]{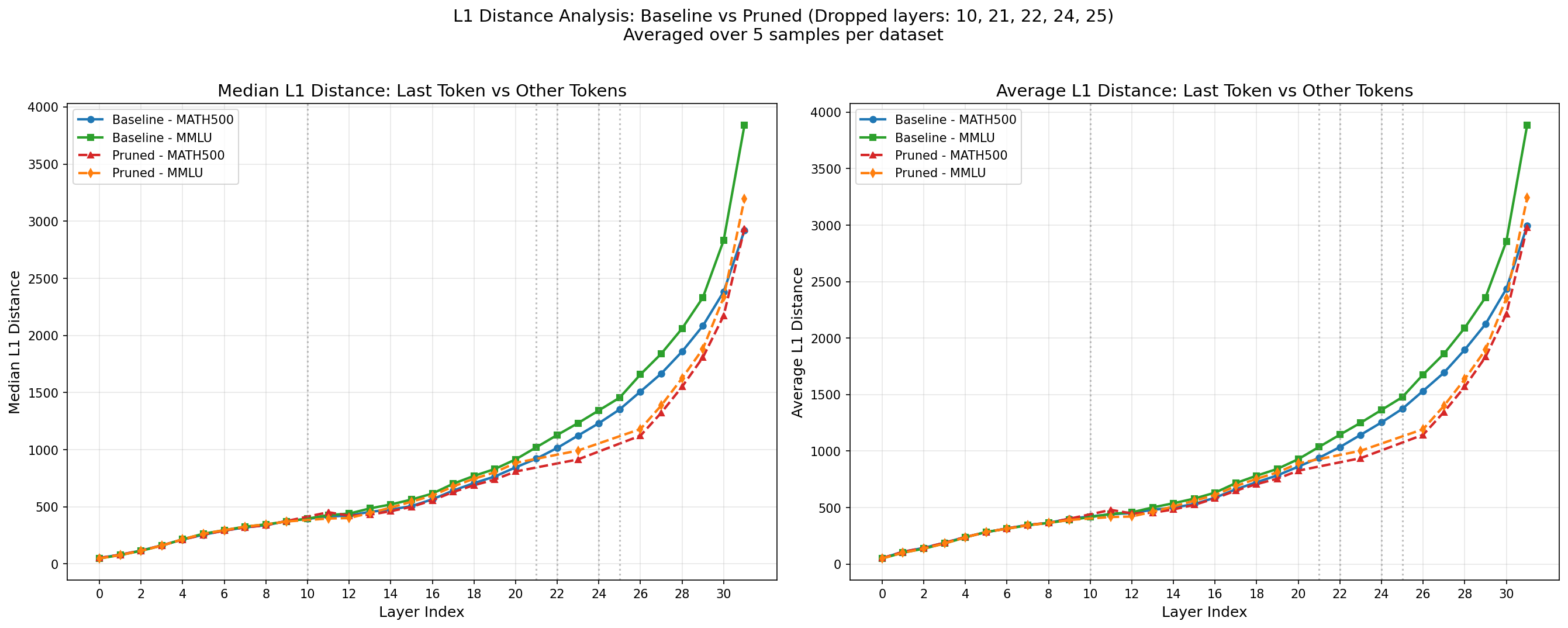}
\caption{The $L_1$ analogue of the $L_2$ analysis in Figure~\ref{fig:pruning_geometry}(b) reproduces the same mechanism: baseline MMLU exhibits amplified token-trajectory distances well above MATH500 from layer 14 onward, and pruning the same five layers sharply reduces that divergence. Agreement between $L_1$ and $L_2$ confirms that the effect is geometric rather than metric-specific, and that \method consistently contracts OOD representational inflation across norms. This cross-metric consistency strengthens the causal link between pruning, norm alignment, and robustness.}

\label{fig:math-mmlu-l1}
\end{figure}
\newpage

\newpage
\clearpage\newpage
\section*{NeurIPS Paper Checklist}

\begin{enumerate}

\item {\bf Claims}
    \item[] Question: Do the main claims made in the abstract and introduction accurately reflect the paper's contributions and scope?
    \item[] Answer: \answerYes{}
    \item[] Justification: The abstract and introduction state the paper's main empirical and mechanistic claims: task-aware pruning improves OOD performance but not ID performance, and this effect is explained through changes in representation geometry. These claims are supported by the controlled regression experiments, fine-tuned LLM experiments, and geometric analyses in Sections~\ref{sec:ood} and~\ref{sec:analysis}.

\item {\bf Limitations}
    \item[] Question: Does the paper discuss the limitations of the work performed by the authors?
    \item[] Answer: \answerYes{}
    \item[] Justification: The paper includes a separate ``Limitations'' section. This section discusses the operational nature of the ID/OOD distinction for pretrained and fine-tuned language models, the fact that the geometric analysis relies on a limited set of representation statistics, the approximate and post hoc nature of the causal surrogate interventions, and the experiments across model scales, task families, architectures, and evaluation settings.

\item {\bf Theory assumptions and proofs}
    \item[] Question: For each theoretical result, does the paper provide the full set of assumptions and a complete (and correct) proof?
    \item[] Answer: \answerNA{}
    \item[] Justification: The paper provides a geometric interpretation and formal notation in Appendix~\ref{appendix:math}, but it does not present formal theoretical results requiring complete proofs. The central contributions are empirical and mechanistic rather than theorem-proving.

\item {\bf Experimental result reproducibility}
    \item[] Question: Does the paper fully disclose all the information needed to reproduce the main experimental results of the paper to the extent that it affects the main claims and/or conclusions of the paper (regardless of whether the code and data are provided or not)?
    \item[] Answer: \answerYes{}
    \item[] Justification: The paper specifies the controlled regression task, model architecture, data-generating distributions, validation distributions, pruning procedure, evaluation protocol, and main LLM benchmarks. Additional training and evaluation details are provided in the appendices, including hyperparameters and evaluation seeds for the controlled experiments.

\item {\bf Open access to data and code}
    \item[] Question: Does the paper provide open access to the data and code, with sufficient instructions to faithfully reproduce the main experimental results, as described in supplemental material?
    \item[] Answer: \answerYes{}
    \item[] Justification: Yes , we do provide github repo containing all the required codes and information available for reproducing the results mentioned in the paper.
    

\item {\bf Experimental setting/details}
    \item[] Question: Does the paper specify all the training and test details (e.g., data splits, hyperparameters, how they were chosen, type of optimizer) necessary to understand the results?
    \item[] Answer: \answerYes{}
    \item[] Justification: The paper specifies the controlled regression setup, the base and shifted coefficient distributions, the transformer architecture, the greedy pruning procedure, the LLM fine-tuning and evaluation tasks, and the representation-extraction procedure. Additional training details, evaluation seeds, and hyperparameters are provided in the appendices.

\item {\bf Experiment statistical significance}
    \item[] Question: Does the paper report error bars suitably and correctly defined or other appropriate information about the statistical significance of the experiments?
    \item[] Answer: \answerYes{}
    \item[] Justification: The main LLM results are reported as mean accuracy with standard deviation across random seeds in Table~\ref{tab:tale-scales}. The controlled experiments also report averages over multiple seeds and fixed evaluation sets in the appendix.

\item {\bf Experiments compute resources}
    \item[] Question: For each experiment, does the paper provide sufficient information on the computer resources (type of compute workers, memory, time of execution) needed to reproduce the experiments?
    \item[] Answer: \answerYes{}
    \item[] Justification: The paper does provide all the compute and experimental details for each experiment in the apendix. 
    
\item {\bf Code of ethics}
    \item[] Question: Does the research conducted in the paper conform, in every respect, with the NeurIPS Code of Ethics \url{https://neurips.cc/public/EthicsGuidelines}?
    \item[] Answer: \answerYes{}
    \item[] Justification: The work studies model pruning and representation geometry using synthetic data, public benchmarks, and existing models. We do not collect private data, conduct human-subject experiments, or release high-risk datasets or models.

\item {\bf Broader impacts}
    \item[] Question: Does the paper discuss both potential positive societal impacts and negative societal impacts of the work performed?
    \item[] Answer: \answerNA{}
    \item[] Justification: The aim of the paper does not deliver any societal impact directly.
    
\item {\bf Safeguards}
    \item[] Question: Does the paper describe safeguards that have been put in place for responsible release of data or models that have a high risk for misuse (e.g., pre-trained language models, image generators, or scraped datasets)?
    \item[] Answer: \answerNA{}
    \item[] Justification: The paper does not introduce or release a new pretrained model, image generator, scraped dataset, or other high-risk asset. It analyzes pruning and representation geometry using synthetic tasks, public benchmarks, and existing models.

\item {\bf Licenses for existing assets}
    \item[] Question: Are the creators or original owners of assets (e.g., code, data, models), used in the paper, properly credited and are the license and terms of use explicitly mentioned and properly respected?
    \item[] Answer: \answerYes{}
    \item[] Justification: The paper cites the datasets, benchmarks, methods, and models used, including MATH500, MMLU, BoolQ, NuminaMath-CoT, Code Alpaca, Llama, and GPT-OSS.

\item {\bf New assets}
    \item[] Question: Are new assets introduced in the paper well documented and is the documentation provided alongside the assets?
    \item[] Answer: \answerNA{}
    \item[] Justification: The paper does not introduce a new dataset, benchmark, or pretrained model as a primary contribution. The controlled regression data are synthetically generated from fully specified distributions.

\item {\bf Crowdsourcing and research with human subjects}
    \item[] Question: For crowdsourcing experiments and research with human subjects, does the paper include the full text of instructions given to participants and screenshots, if applicable, as well as details about compensation (if any)? 
    \item[] Answer: \answerNA{}
    \item[] Justification: The paper does not involve crowdsourcing or research with human subjects.

\item {\bf Institutional review board (IRB) approvals or equivalent for research with human subjects}
    \item[] Question: Does the paper describe potential risks incurred by study participants, whether such risks were disclosed to the subjects, and whether Institutional Review Board (IRB) approvals (or an equivalent approval/review based on the requirements of your country or institution) were obtained?
    \item[] Answer: \answerNA{}
    \item[] Justification: The paper does not involve crowdsourcing or human-subject research, so IRB approval or equivalent review is not applicable.

\item {\bf Declaration of LLM usage}
    \item[] Question: Does the paper describe the usage of LLMs if it is an important, original, or non-standard component of the core methods in this research? Note that if the LLM is used only for writing, editing, or formatting purposes and does \emph{not} impact the core methodology, scientific rigor, or originality of the research, declaration is not required.
    \item[] Answer: \answerNA{}
    \item[] Justification: LLMs are objects of study in the experiments, not tools used as an important, original, or non-standard component of the research methodology. Any use of LLMs for writing, editing, or formatting does not affect the scientific methodology, results, or originality of the work.

\end{enumerate}

\end{document}